\pdfoutput=1

\documentclass[11pt]{article}

\usepackage[preprint]{acl}

\usepackage{times}
\usepackage{latexsym}
\usepackage{lipsum}
\usepackage{booktabs}
\usepackage[labelfont=bf]{caption}
\usepackage{multirow}
\usepackage{graphicx}
\usepackage{amssymb}
\usepackage{amsmath}
\usepackage{soul}
\usepackage{subcaption}
\usepackage{enumitem}
\usepackage{hyperref}

\usepackage[T1]{fontenc}

\usepackage[utf8]{inputenc}

\usepackage{microtype}

\usepackage{inconsolata}

\newcommand*\rot{\rotatebox{90}}
\newcommand{\mistral}[1]{{\tt Mistral-$#1$B}}
\newcommand{\llama}[1]{{\tt LLaMA-$#1$B}}
\newcommand{\llamatwo}[2]{{\tt LLaMA$#1$-$#2$B}}
\newcommand{\vicuna}[1]{{\tt Vicuna-$#1$B}}
\newcommand{\zephyr}[1]{{\tt Zephyr-$#1$B}}
\newcommand{\goodsum}{\textsc{Good-Sum}}
\newcommand{\sbadsum}{\textsc{SBad-Sum}}
\newcommand{\vbadsum}{\textsc{VBad-Sum}}
\newcommand{\trainset}{{\tt train}}
\newcommand{\validset}{{\tt validation}}

\newcommand{\propdata}{\textsc{PromptOpinSumm}}
\newcommand{\maspcov}{{\tt aspect-coverage}}
\newcommand{\mopfaith}{{\tt opinion-faithfulness}}
\newcommand{\mopcov}{{\tt opinion-coverage}}
\newcommand{\mconcise}{{\tt conciseness}}
\newcommand{\mrelevance}{{\tt relevance}}
\newcommand{\mhallucination}{{\tt hallucination}}
\newcommand{\mlangcorr}{{\tt language-correctness}}
\newcommand{\maspcovb}{{\tt \textbf{aspect-coverage}}}
\newcommand{\mopfaithb}{{\tt \textbf{opinion-faithfulness}}}
\newcommand{\mopcovb}{{\tt \textbf{opinion-coverage}}}
\newcommand{\mconciseb}{{\tt \textbf{conciseness}}}
\newcommand{\mrelevanceb}{{\tt \textbf{relevance}}}
\newcommand{\mhallucinationb}{{\tt \textbf{hallucination}}}
\newcommand{\mlangcorrb}{{\tt \textbf{language-correctness}}}
\newcommand{\supmodel}{\textsc{Supervised}}
\newcommand{\naivemeanmodel}{\textsc{NaiveMean}}
\newcommand{\syntheticmodel}{\textsc{Synth-Feedback}}
\newcommand{\inductivebiasmodel}{\textsc{Inductive-Bias}}
\newcommand{\rmop}{$\varphi_{op}$}
\newcommand{\elo}{\textit{Elo}}
\newcommand{\rouge}[1]{\textsc{Rouge}-$#1$}
\newcommand{\gpt}[1]{\textsc{Gpt}-$#1$}
\newcommand{\tjmodel}{\textsc{Op-Sum-Gen}}
\newcommand{\dpomodel}{\textsc{DPO}}
\newcommand{\medosmodel}{\textsc{Medos}}
\newcommand{\scenei}{\textsc{Scene-I}}
\newcommand{\sceneii}{\textsc{Scene-II}}
\newcommand{\sceneiii}{\textsc{Scene-III}}
\newcommand{\sceneiv}{\textsc{Scene-IV}}
\newcommand{\opinpref}{\textsc{OpinPref}}
\newcommand{\amazonb}{\textsc{Amazon}}
\newcommand{\amazonbr}{\textsc{Amazon-R}}
\newcommand{\amazonbrdq}{\textsc{Amazon-Rdq}}
\newcommand{\oposumb}{\textsc{Oposum+}}
\newcommand{\oposumbr}{\textsc{Oposum-R}}
\newcommand{\oposumbrdq}{\textsc{Oposum-Rdq}}
\newcommand{\flipkartb}{\textsc{Flipkart}}
\newcommand{\flipkartbr}{\textsc{Flipkart-R}}
\newcommand{\flipkartbrdq}{\textsc{Flipkart-Rdq}}
\newcommand{\hyposhort}{\textit{infusing domain knowledge into} $\varphi$ \textit{to reap benefits of RLHF with modest human preference data}}

\newlist{MyIndentedList}{itemize}{4}
\setlist[MyIndentedList,1]{%
    label={},
    noitemsep,
    leftmargin=0pt,
    }
\setlist[MyIndentedList]{%
    label={},
    noitemsep,
    }
\newcommand\blfootnote[1]{%
  \begingroup
  \renewcommand\thefootnote{}\footnote{#1}%
  \addtocounter{footnote}{-1}%
  \endgroup
}

\title{Leveraging Domain Knowledge for Efficient Reward Modeling in RLHF: A Case-Study in E-Commerce Opinion Summarization}

\author{Swaroop Nath$^\dag$$^\clubsuit$,
Tejpalsingh Siledar$^\dag$$^\clubsuit$, Sri Raghava$^\dag$$^\clubsuit$, Rupasai Rangaraju$^\dag$$^\clubsuit$, 
\\ 
\textbf{Harshad Khadilkar$^\clubsuit$, Pushpak Bhattacharyya$^\clubsuit$,} \\
\textbf{Suman Banerjee$^\heartsuit$, Amey Patil$^\heartsuit$, Sudhanshu Shekhar Singh$^\heartsuit$, }\\
\textbf{Muthusamy Chelliah$^\heartsuit$, Nikesh Garera$^\heartsuit$}\\
        $^\clubsuit$Computer Science and Engineering, IIT Bombay, India \\
        $^\heartsuit$Flipkart, India \\
        {\tt swaroopnath}{\tt @cse.iitb.ac.in}
        }

\begin{document}
\maketitle
\blfootnote{$^\dag$ Equal contribution}
\begin{abstract}
Reinforcement Learning from Human Feedback (RLHF) has become a dominating strategy in aligning Language Models (LMs) with human values/goals. The key to the strategy is learning a reward model ($\varphi$), which can reflect the latent reward model of humans. While this strategy has proven effective, the training methodology requires a lot of human preference annotation (usually in the order of tens of thousands) to train $\varphi$. Such a large-scale annotation is justifiable when it's a one-time effort, and the reward model is universally applicable. However, human goals are subjective and depend on the task, requiring task-specific preference annotations, which can be impractical to fulfill. To address this challenge, we propose a novel approach to infuse domain knowledge into $\varphi$, which reduces the amount of preference annotation required ($21\times$), omits Alignment Tax, and provides some interpretability. We validate our approach in E-Commerce Opinion Summarization, with a significant reduction in dataset size (to just $940$ samples) while advancing the SOTA ($\sim4$ point \rouge{L} improvement, $68\%$ of times preferred by humans over SOTA). Our contributions include a novel Reward Modeling technique and two new datasets: \propdata \:(supervised data for Opinion Summarization) and \opinpref \:(a gold-standard human preference dataset). The proposed methodology opens up avenues for efficient RLHF, making it more adaptable to applications with varying human values. We release the artifacts\footnote{Code: \href{https://github.com/swaroop-nath/reward-approx-social-choice-opp-summ}{\tt github.com/efficient-rlhf}. \textsc{PromptOpinSumm}: \href{https://huggingface.co/datasets/swaroop-nath/prompt-opin-summ}{\tt hf.co/prompt-opin-summ}. \textsc{OpinPref}: \href{https://huggingface.co/datasets/swaroop-nath/opin-pref}{\tt hf.co/opin-pref}} for usage under MIT License.
\end{abstract}

\section{Introduction}\label{sec:introduction}

Reinforcement Learning from Human Feedback (RLHF) \cite{ziegler-etal-2019-fine-tuning-lms-human-pref, ouyang-etal-2022-instruct-gpt} is a prominent approach in aligning Language Models (LMs) with human values. Human values are represented by a function ($\varphi$), which ultimately acts as the reward in the RLHF training. For an output $Y$ ($=y_1, y_2, \cdots, y_n$) to some input $X$ ($=x_1, x_2, \cdots, x_m$), $\varphi$ performs the mapping $(X, Y) \rightarrow r$. The reward function $\varphi$ is latent to humans and manifests in human preferences. Preference Modeling techniques, such as Bradley-Terry model \cite{bradley-terry-human-pref-model}, Plackett-Luce models \cite{plackett-human-pref-model, luce-human-pref-model} are used to learn $\varphi$ from preference data, of the form: $\mathcal{D} = \{(X, Y_w, Y_l) \> | \> Y_w \succ Y_l\}$\footnote{$Y_w \succ Y_l$, in this entire paper, signifies that the output $Y_w$ is preferred over the output $Y_l$; $w$: \textit{win}, $l$: \textit{loss}.}.

\begin{figure}[t]
    \centering
    \includegraphics[width=\linewidth]{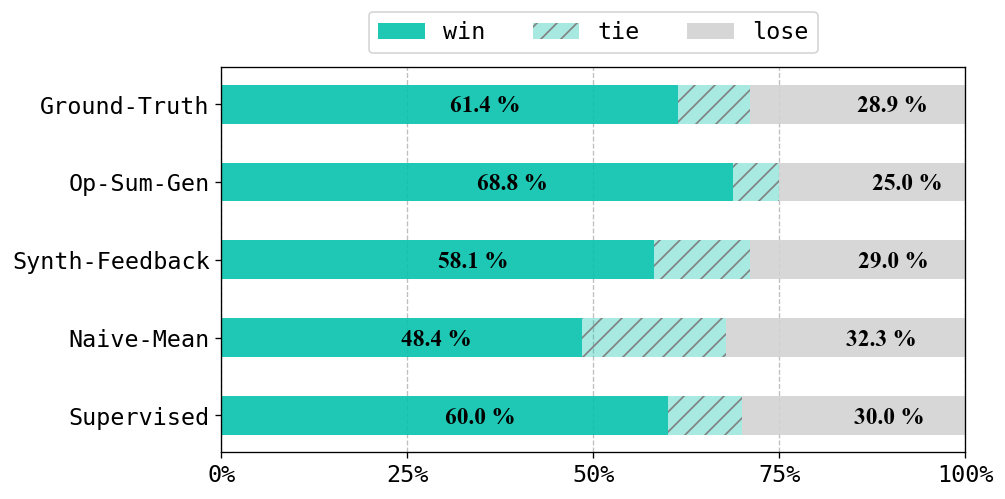}
    \caption{Human Eval: Pairwise win-tie-loss percentage of \inductivebiasmodel \:model (our proposed model) vs. ground truth summary and summary from other models, for \amazonb \:benchmark. We see that our proposed approach (\hyposhort) helps \inductivebiasmodel \:model achieve summaries which are always preferred (Section \ref{subsec:eval-results}).}
    \label{fig:human-eval-bar-chart}
\end{figure}

In contemporary works \cite{ziegler-etal-2019-fine-tuning-lms-human-pref, bai-etal-2022-hh-rlhf-data, ouyang-etal-2022-instruct-gpt, rafailov-etal-2023-dpo}, the reward functions are Large LMs (LLMs; pretrained Transformers) themselves. The text data, ($X$, $Y_w$) and ($X$, $Y_l$) are directly fed to $\varphi$, for training. Such a formulation necessitates large-scale human preference data to train the LLM (millions/billions of parameters). Typically the size of $\mathcal{D}$ varies from $20K$ \cite{nakano-etal-2021-webgpt, bai-etal-2022-hh-rlhf-data} to $> 200K$ \cite{chaudhari-etal-2022-shp-data}. Such a large-scale annotation is justifiable when it's a one-time effort, and the trained $\varphi$ is universally applicable, irrespective of the nature of the downstream task. However, human values are subjective \cite{jiang-etal-2022-delphi, sorensen-etal-2023-value-pluralism}. For instance, \textit{hallucination would be desired in Creative Writing}, \textit{but not in Question-Answering}. This means that \textbf{depending on the downstream task, the reward function} $\mathbf{\varphi}$ \textbf{must have varying characteristics}. Collecting human preferences for all such tasks is impractical.

Motivated to resolve this need, we propose a novel reward modeling methodology, significantly reducing preference data requirements. We draw on the insight that $\varphi$ is dependent on the downstream task and, hence, can utilize its task/domain\footnote{We use task and domain interchangeably in the paper.} knowledge. Specifically, \textit{$\varphi$ lies in a low-dimensional manifold, whose dimensions can be deduced using domain knowledge}. Such an inductive bias \textit{reduces the number of samples}\footnote{An example: For a function, $f: (x_1, x_2, x_3, \cdots, x_m) \rightarrow y$, assuming that $f$ is a linear combination of $x_i$ (Linear Regression) reduces the training data requirement. Assuming no functional form (Feed-Forward Neural Network) would require more data.} needed to train $\varphi$. Concretely, our \textbf{hypothesis} is: \textit{An inductive bias infused $\varphi$ \underline{can help achieve alignment} with human values for a task, with \underline{modest human preference annotations}}. Specifically, we say that $\varphi_{\tau}$ (reward model for a domain $\tau$) can be modelled by some numeric features $v_1, v_2, \cdots, v_n$. These $n$ features fully characterize\footnote{Example of such characterization: Features like \textit{fluency}, \textit{coherence}, etc. can characterize text generated by an LLM.} the outputs from the LLM on some input. Thus, instead of training $\varphi_\tau$ on the text data ($\{(X, Y_w, Y_l) \>|\> Y_w \succ Y_l\}$), we use the $n$ features. Such a formulation for $\varphi$ \textbf{brings interpretability}---which features influence human preference the most (Section \ref{sec:analysis}), and is \textbf{free from Alignment Tax} (degradation of language capabilities of an LLM post reward modeling; \citet{bai-etal-2022-hh-rlhf-data}) as we do not use an LLM to model $\varphi$.

We experimentally prove our hypothesis in the domain of E-Commerce Opinion Summarization \cite{brazinskas-etal-2020-unsupervised, amplayo-etal-2021-aspect, siledar-etal-2023-opinion-summ}---the task of summarizing user reviews for a product. In addition to advancing SOTA, we also analyze how our approach helps the model achieve alignment with human values for Opinion Summarization (Section \ref{sec:analysis}).

Our contributions are:

\begin{enumerate}
    \item A novel Reward Modeling technique for RLHF, which leverages Domain Knowledge to achieve alignment with human values while significantly reducing human preference annotation. In the domain of Opinion Summarization, we achieve alignment while reducing\footnote{As compared to the smallest publicly available preference data. The smallest publicly available preference data is not in the domain of Opinion Summarization.} the dataset size by $>\mathbf{21\times}$. Our approach advances SOTA: at least $\sim\mathbf{4}$-point \rouge{L} improvement (Tables \ref{tab:automatic-eval-amazon}, \ref{tab:automatic-eval-flipkart} and \ref{tab:automatic-eval-oposum}; Section \ref{subsec:eval-results}), and humans prefer our models' outputs $\mathbf{68}\%$ over SOTA (Figure \ref{fig:human-eval-bar-chart}; Section \ref{subsec:eval-results}).
    \item Two new datasets: \propdata \:and \opinpref. \propdata \:includes reviews and summaries for $25763$ products ($229521$ summaries), for training and validation. \opinpref \:is a gold-standard human preference dataset (with $940$ instances) in the domain of Opinion Summarization.
\end{enumerate}

\section{Related Works}\label{sec:related-works}

\textbf{Steering Language Models (LMs) towards human goals}: Steering LMs towards human goals/values refers to the task of training LMs to generate text which is more aligned with human values, such as `\textit{text should not have harmful content}', `\textit{it should be polite}', etc. Such a task necessitates a human presence in the training loop of these LMs. In recent times, Reinforcement Learning from Human Feedback (RLHF) \cite{ziegler-etal-2019-fine-tuning-lms-human-pref, askell-etal-2021-general, bai-etal-2022-hh-rlhf-data, ouyang-etal-2022-instruct-gpt, liu-etal-2022-aligning} has emerged as an effective solution---by incorporating Reward Models, which reflect latent reward models within humans, into the training pipeline. These reward models are trained on human preference datasets \cite{ziegler-etal-2019-fine-tuning-lms-human-pref, nakano-etal-2021-webgpt, chaudhari-etal-2022-shp-data}, which are typically of the order of tens of thousands, in size. Dependence on high-quality, large-sized preference data is an obstacle for RLHF.

Recently, Reinforcement Learning from AI Feedback (RLAIF) \cite{bai-etal-2022-constitutional-ai, kim-etal-2023-aligning, lee-etal-2023-rlaif} has emerged as an alternative. It attempts to reduce the dependence on human preference datasets by using Large LMs (LLMs) as preference data generators. While this is a scalable approach to steering LMs, there is no guarantee that the preference dataset generated by LLMs reflects human goals. In our work, we propose a different solution, which promises to use human preference data but provides a way to reduce the required size drastically. To the best of our knowledge, we are the first to attempt this.

\noindent\textbf{Opinion Summarization}: Opinion Summarization \cite{hu-liu-2004-opinion-mining, brazinskas-etal-2020-unsupervised, amplayo-etal-2021-aspect, siledar-etal-2023-opinion-summ} is the task of summarizing user reviews. Specifically, we look at E-Commerce Opinion Summarization, where user reviews are on products. These reviews contain aspects of the product and users' sentiments/opinions towards those aspects. Previous works \cite{brazinskas-etal-2020-unsupervised, siledar-etal-2023-synthesize} in E-Commerce Opinion Summarization have used \textit{Self-Supervised} training methodology. In this context, self-supervision refers to picking one of the $N$ available reviews as a summary, commonly called \textit{pseudo-summary}, and training the model on the remaining $N-1$ reviews to generate the pseudo-summary. The theme of solutions \cite{chu-etal-2019-mean-sum, brazinskas-etal-2020-unsupervised, siledar-etal-2023-opinion-summ, siledar-etal-2023-synthesize} have mostly centered around Supervised Learning. The core problem has always been getting good synthetic datasets for training. More recently, Prompting \cite{bhaskar-etal-2023-prompted} has been explored to solve the task. \citet{bhaskar-etal-2023-prompted} move away from making a better synthetic dataset generation pipeline and test \gpt{3.5} for Opinion Summarization.

We \underline{do not} propose a new synthetic dataset generation methodology. Rather, we generate training data using an open-source LLM (\mistral{7}), to test our hypothesis. To the best of our knowledge, we are the first to propose such a dataset for training Opinion Summarizers. Such an approach has been explored for Generic Text Summarization \cite{wang-etal-2023-self-instruct, taori-etal-2023-alpaca, peng-etal-2023-instruction}. \citet{taori-etal-2023-alpaca} fine-tune \llama{7} \cite{touvron-etal-2023-llama} using Instruction-Tuning dataset generated using \gpt{3}. \citet{peng-etal-2023-instruction} fine-tune \llama{7} using a dataset generated by \gpt{4}.

\section{Dataset}\label{sec:dataset}

Previous works \cite{brazinskas-etal-2020-unsupervised, siledar-etal-2023-synthesize} in Opinion Summarization have used \textit{Self-Supervised} training methodology, where $N-1$ reviews are used as input, and the left out review is used as a pseudo-summary (Section \ref{sec:related-works}). Although these self-supervision datasets have helped further Opinion Summarization research, the approach has several shortcomings: the summaries always present a one-person rather than the consensus view, the summaries are reviews and might not have good coverage of aspects and opinions, etc. We move away from self-supervision to overcome these shortcomings and propose a new dataset. In the rest of this Section, we describe (\textit{a}) \propdata: a new dataset to train Opinion Summarizers, (\textit{b}) the benchmarks we used for evaluation, and (\textit{c}) \opinpref: gold-standard preference dataset for Opinion Summarization.

\subsection{\propdata \:Dataset}\label{subsec:prompt-opin-summ-dataset}
We prompt the instruction-tuned \mistral{7} model \cite{jiang-etal-2023-mistral} to generate an opinion summary given product reviews. We also tried other open-source LLMs available at the time of the work, such as \llamatwo{2}{7}, \llamatwo{2}{13} \cite{touvron-etal-2023-llama-two}, \vicuna{7}, \vicuna{13} \cite{chiang-etal-2023-vicuna}, \zephyr{7} \cite{tunstall-etal-2023-zephyr}. However, we found that \mistral{7} leads to better summaries. We limit ourselves to open-source models due to cost. Appendix \ref{asec:generated-data-appendix} includes examples and qualitative analysis. We use the Amazon dataset \cite{he-mcaulay-2016-amazon}, which has reviews for $\sim180k$ products. We randomly sample reviews for $20763$ products for \trainset \:set and $5000$ products for \validset \:set. Specifically, we prompt the model to generate opinion summaries of $3$ different qualities: Good (codenamed \goodsum), Slightly Bad (codenamed \sbadsum), and Very Bad (codenamed \vbadsum). We generate multiple opinion summaries ($3$ at most) per quality. We provide reasoning for generating multiple summaries of different qualities in the extended discussion of our approach (Appendix \ref{asec:rlhf-training}). We generate $184620$ summaries for \trainset \:set and $44901$ summaries for \validset \:set (see Appendix \ref{asec:generated-data-appendix}).

\subsection{Benchmarks for Evaluation}\label{subsec:benchmark-datasets}

We use $9$ Opinion Summarization benchmarks for evaluation. $3$ of these benchmarks are the Amazon test set (\citet{brazinskas-etal-2020-unsupervised}, codenamed \amazonb), the Oposum+ test set (\citet{amplayo-etal-2021-aspect}, codenamed \oposumb) and the Flipkart test set (\citet{siledar-etal-2023-opinion-summ}, codenamed \flipkartb). \amazonb \:has reviews for $32$ products from $4$ domains, \oposumb \:has reviews for $60$ products from $6$ domains and \flipkartb \:has reviews for $147$ products from $3$ domains.

Although these $3$ benchmarks have been used widely, they have several shortcomings. For instance, \amazonb \:was developed by asking annotators to write a summary in first-person point-of-view. This causes problems such as summaries seeming personal rather than consensus opinions (which can include mixed sentiment), incomplete coverage of aspects and opinions, etc. Thus, using such \textit{pseudo-summaries} for reference-based evaluations (\textsc{Rouge}, \textsc{BertScore}) on such a benchmark is not a correct portrayal of the models' performances. We highlight the shortcomings in detail in Appendix \ref{asec:benchmark-problems}. \citet{siledar-etal-2024-product} recently provided $6$ new benchmarks (\amazonbr, \amazonbrdq, \oposumbr, \oposumbrdq, \flipkartbr, \flipkartbrdq) which are revamped versions (by getting rid of the shortcomings) of the aforementioned $3$ benchmarks. We primarily rely on these $6$ for our conclusions. Appendix \ref{asec:benchmark-problems} includes more details on domains and summary statistics.

\subsection{\opinpref \:Dataset}
We create \opinpref \:by asking humans to rank opinion summaries for given reviews. We utilize domain experts (annotator details in Appendix \ref{asec:annotator-details}) to perform the annotation. We believe that aligning with the internal reward model of domain experts would lead to better opinion summaries. We provide the domain expert with product reviews and two opinion summaries (products are sampled from the \propdata \:dataset). The domain expert notifies which of the two summaries they prefer. We use this to construct a dataset of the form: $\mathcal{D}_h = \{(R, s_w, s_l) \> | \> s_w \succ s_l\}$, where $R$ is the set of reviews and $s_w$ and $s_l$ are opinion summaries. We construct a dataset of $940$ samples. We observe a Fleiss' Kappa ($\kappa$) score of $62.67\%$ (substantial aggrement; aggrement is substantial when $60\% \leq \kappa < 80\%$). Appendix \ref{asec:opinpref-stats} includes statistics on the dataset.

\begin{figure*}
    \centering
    \includegraphics[width=0.95\linewidth]{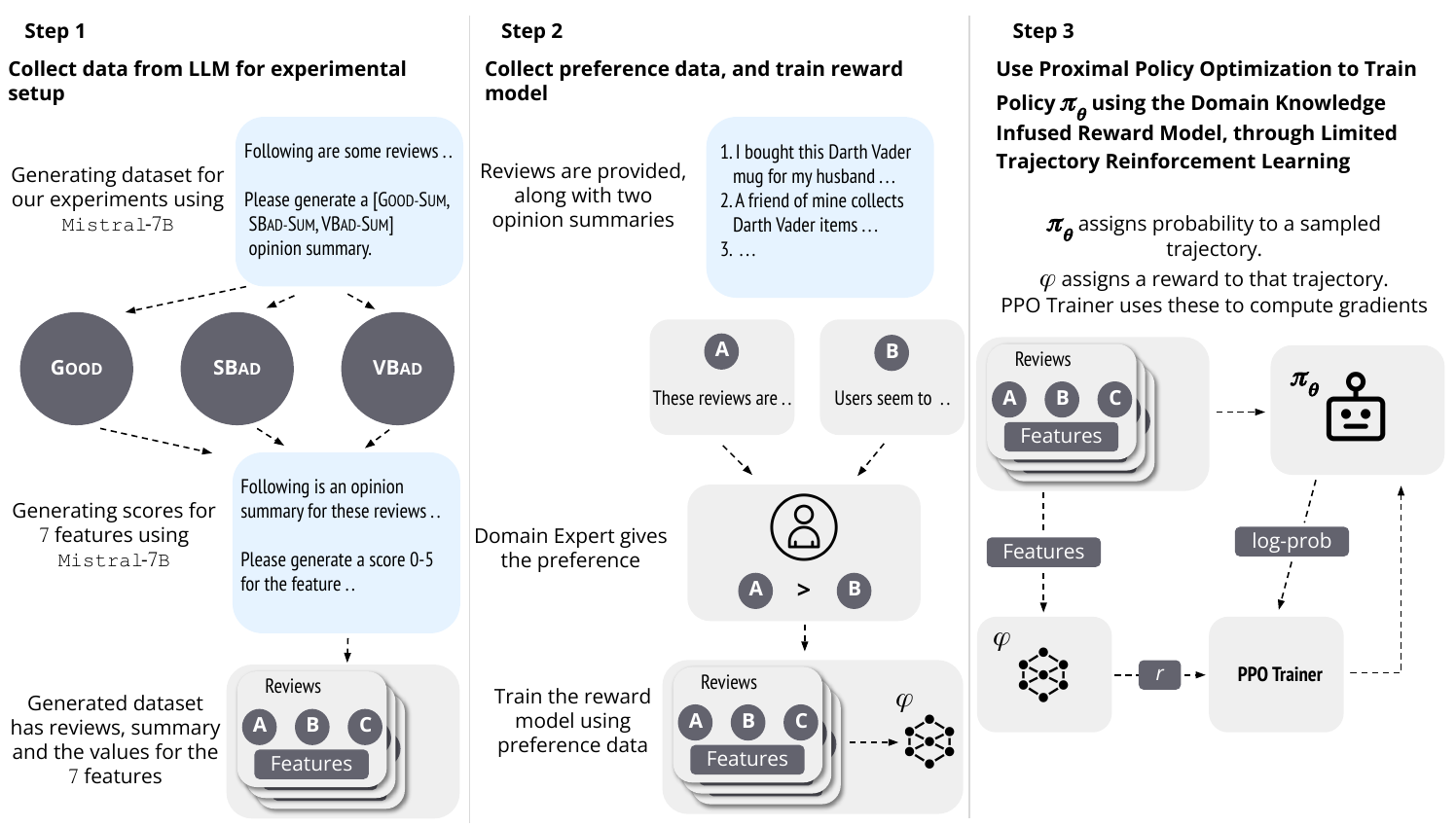}
    \caption{Overview of our approach. Step-$1$: We generate a new dataset for training Opinion Summarizers: \propdata, by prompting \mistral{7} \:model. Again, we use \mistral{7} \:to compute values for the $7$ features discussed in Section \ref{subsec:reward-modelling}. Step-$2$: We ask humans (domain experts) for their preference, given reviews and two opinion summaries (A, B). We use the preference data and the features to train the reward model, \rmop. Step-$3$: We sample instances from \propdata \:dataset; \rmop \:assigns a score to the sampled summaries, the policy, $\pi_\theta$, assigns \textit{log probabilities} to these summaries. Proximal Policy Optimization uses these to update $\pi_\theta$.}
    \label{fig:rlhf-pipeline}
\end{figure*}

\section{Efficient Reward Modeling}\label{sec:technique}

We highlighted in Section \ref{sec:introduction} how the reward model ($\varphi$) can depend on the downstream task. Such dependence necessitates task/domain-specific human preference datasets, which are costly and time-consuming to create. This creates an obstacle in employing RLHF in task/domain-specific setups, thus hindering the steering of LLMs towards task/domain-specific human values.

We solve this challenge by leveraging domain knowledge. \textbf{The key insight is that we can use the domain knowledge to impart some inductive biases into the mathematical modeling of} $\mathbf{\varphi}$. This would significantly reduce the amount of data required for training $\varphi$. Specifically, we say that $\varphi_{\tau}$ (reward model for a domain $\tau$) can be modelled by some numeric features $v_1, v_2, \cdots, v_n$. These $n$ features fully characterize\footnote{Example of such characterization: Features like \textit{fluency}, \textit{coherence}, etc. can characterize text generated by an LLM.} the outputs from the LLM on some input. Thus, instead of training $\varphi_\tau$ on the text data ($\{(X, Y_w, Y_l) \>|\> Y_w \succ Y_l\}$), we use the $n$ features. Such a formulation for $\varphi$ also brings interpretability and frees $\varphi$ from Alignment Tax.

In Section \ref{subsec:reward-modelling}, we detail our technique for the task/domain of E-Commerce Opinion Summarization---the task of summarizing user reviews for a product. Typically, user reviews discuss several aspects of a product and opinions/sentiments towards these aspects. An opinion summary must reflect all the aspects discussed by the input reviews and the opinions expressed towards these aspects. We discuss how we leverage such desirable properties to model $\varphi$.

\subsection{Inducing Domain Knowledge}\label{subsec:reward-modelling}

We identify desirable properties in an opinion summary with the help of domain experts\footnote{Domain experts are from an E-Commerce platform.}. We held multiple discussions to finalize the set of desirable properties. We show that \textbf{these properties are correlated to humans' judgement of summary} in Appendix \ref{asec:reward-modelling-featues} (Table \ref{tab:feature-correlation}). Based on these properties, we model \rmop \:(reward model for opinion summarization) as: \rmop $\>= f(v)$, where $v \in$ \{\maspcov, \mopfaith, \mopcov, \mconcise, \mrelevance, \mhallucination, \mlangcorr\}. The features \maspcov, \mopfaith$\>$ and \mopcov$\>$ check if the generated opinion summary covers all mentioned aspects and opinions faithfully. The features \mconcise, \mrelevance$\>$, and \mhallucination$\>$ check if the generated summary is concise, relevant to the input reviews, and is free from hallucination. The feature \mlangcorr$\>$ checks if the generated text follows the language rules. We provide more details in Appendix \ref{asec:reward-modelling-featues}. These features, together, characterize the goodness of an opinion summary. We instruct \mistral{7} (Appendix \ref{asec:reward-modelling-featues}) to generate values for these features for an opinion summary, given reviews. We denote this transformation (from reviews and summary to $7$ features) using $\Phi$.

We train \rmop \:using \opinpref, which is of the form: $\mathcal{D}_h = \{(R, s_w, s_l) \> | \> s_w \succ s_l\}$, where $R$ is the set of reviews and $s_w$ and $s_l$ are opinion summaries.  We parameterize \rmop \:using a Feed-Forward Neural Network and train it using the \elo-loss \cite{ouyang-etal-2022-instruct-gpt, glaese-etal-2022-sparrow} (Equation \ref{eqn:elo-loss}; $\Phi(R, s_i)$ uses \mistral{7} to compute the $7$ features; only \rmop \:is trainable, $\Phi$ is not).

After such an efficient reward modeling, we use \rmop \:for regular RLHF training (Appendix \ref{asec:rlhf-training}) to get an Opinion Summarizer aligned with human goals. We illustrate the whole flow in Figure \ref{fig:rlhf-pipeline}.

\vspace{-1.5em}
\begin{align}\label{eqn:elo-loss}
    \mathcal{L}_{pr} = - \mathbb{E}_{(R, s_w, s_l) \sim \mathcal{D}_h} \Big[&\log \sigma \big(\varphi_{op}(\Phi(R, s_l)) \nonumber \\
    &- \varphi_{op}(\Phi(R, s_w)) \big) \Big]
\end{align}


\section{Experiments}

We test our technique against the State-of-the-Art (SOTA) models, and strong Reinforcement Learning (RL) and RLHF baselines (our design and contemporary works). We list the questions we attempt to answer (through the experiments) in Section \ref{subsec:models-list}. We conduct automatic, human, and \gpt{4} evaluations to verify our claim. We find that our proposed technique excels significantly. In the rest of the section, we describe our models (Section \ref{subsec:models-list}) and evaluation results (Section \ref{subsec:eval-results}).

\subsection{Models \& Objectives}\label{subsec:models-list}
We train the following models:

\noindent\textbf{\supmodel}: This is a supervised model trained using Maximum Likelihood Estimation.

\noindent\textbf{\naivemeanmodel}: This is a Reinforcement Learning model, where the reward is computed by averaging the feature values obtained using $\Phi$.

\noindent\textbf{\syntheticmodel}: This is a Reinforcement Learning from Synthetic Feedback (RLSF) \cite{kim-etal-2023-aligning} model. For this, we use a reward model trained on the implicit preference \goodsum \:$\succ$ \sbadsum \:$\succ$ \vbadsum. \citet{kim-etal-2023-aligning} show that RLSF is an effective surrogate for RLHF when no human preference data is available. We train this reward model using Equation \ref{eqn:elo-loss} too.

\noindent\textbf{\inductivebiasmodel}: This RLHF model is trained following our hypothesis (\hyposhort). We train \rmop \:using \opinpref \:dataset.

With these models, we ask the following questions in our experiments:

\noindent\textbf{\scenei}: How effective is our technique (\hyposhort) over and above the usage of a good training dataset? A comparative evaluation of \supmodel \:and \inductivebiasmodel \:would answer this.

\noindent\textbf{\sceneii}: How effective is our technique over and above vanilla RL? A comparative evaluation of \naivemeanmodel \:and \inductivebiasmodel \:would answer this.

\noindent\textbf{\sceneiii}: How effective is our technique over contemporary RLHF techniques, which work without preference data? A comparative evaluation of \syntheticmodel \:and \inductivebiasmodel \:would answer this.

\noindent\textbf{\sceneiv}: How effective is our technique, agnostic of the preference data? This question is raised to answer whether the gains are solely due to the good quality of \opinpref, or the approach. A comparative evaluation between \dpomodel \:(\citet{rafailov-etal-2023-dpo}, which uses \opinpref \:in a supervised fashion) and \inductivebiasmodel \:would answer this.


In addition to the above questions, we also check how our models fare against the SOTA (\tjmodel: \citet{siledar-etal-2023-synthesize}, \medosmodel: \citet{siledar-etal-2024-product}, etc.). We \textit{do not} use vanilla RLHF \cite{ziegler-etal-2019-fine-tuning-lms-human-pref, bai-etal-2022-hh-rlhf-data} as a baseline, as it requires huge human preference data. Given that the goal of the paper is not to propose a new RLHF technique, but rather to propose a way to use RLHF with modest human preference annotations, omitting vanilla RLHF as a baseline does not affect our conclusions in any way.

We use BART-Large \cite{lewis-etal-2020-bart} for all of our models. The choice of the model is governed by two factors: (\textit{a}) It provides a similar environment (model size) for comparison with SOTA, (\textit{b}) We find that LLMs (\mistral{7}, \llamatwo{2}{7}, \zephyr{7}, etc.) are already quite good at opinion summarization; thus any performance benefits (over SOTA) in those models cannot be reliably attributed to our approach. We include implementation details in Appendix \ref{asec:implementation-details}.

\begin{table*}[t]
    \centering
    \resizebox{2\columnwidth}{!}{
        \begin{tabular}{*{13}c}
            \toprule
            & \multirow{2}{*}{Model-Code} & \multicolumn{3}{c}{\amazonb} && \multicolumn{3}{c}{\amazonbr} && \multicolumn{3}{c}{\amazonbrdq} \\
            \cmidrule{3-5}\cmidrule{7-9}\cmidrule{11-13}
            && R-$1 \uparrow$ & R-$2 \> \uparrow$ & R-L$\> \uparrow$ && R-$1 \> \uparrow$ & R-$2 \> \uparrow$ & R-L$\> \uparrow$ && R-$1 \> \uparrow$ & R-$2 \> \uparrow$ & R-L$\> \uparrow$ \\
            \midrule
            \multirow{6}{*}{\rot{\textit{Prior Works}}} & MeanSum \cite{chu-etal-2019-mean-sum} & $29.20$ & $4.70$ & $18.15$ && $-$ & $-$ & $-$ && $-$ & $-$ & $-$ \\
            & CopyCat \cite{brazinskas-etal-2020-unsupervised} & $31.97$ & $5.81$ & $20.16$ && $20.09$ & $1.79$ & $12.94$ && $20.54$ & $1.94$ & $13.85$ \\
            & PlanSum \cite{amplayo-lapata-2020-unsupervised} & $32.87$ & $6.12$ & $19.05$ && $20.49$ & $1.76$ & $12.44$ && $19.09$ & $1.58$ & $12.02$ \\
            & MultimodalSum \cite{im-etal-2021-self} & $34.19$ & $7.05$ & $20.81$ && $21.43$ & $1.58$ & $13.20$ && $20.39$ & $2.08$ & $12.83$ \\
            & \tjmodel \:\cite{siledar-etal-2023-synthesize} & $\mathbf{35.46}$ & $\underline{7.30}$ & $\mathbf{21.50}$ && $-$ & $-$ & $-$ && $-$ & $-$ & $-$ \\
            & \medosmodel \:\cite{siledar-etal-2024-product} & $\underline{34.63}$ & $\mathbf{7.48}$ & $\underline{20.97}$ && $23.92$ & $2.27$ & $14.69$ && $25.44$ & $4.16$ & $16.45$ \\
            \midrule
            \multirow{5}{*}{\rot{\textit{Ours'}}} 
            & \dpomodel & $23.96$ & $4.54$ & $14.27$ && $26.37$ & $4.25$ & $15.03$ && $25.13$ & $3.84$ & $14.86$ \\
            & \supmodel & $28.99$ & $4.90$ & $16.91$ && $32.52$ & $5.96$ & $18.07$ && $30.46$ & $\underline{5.49}$ & $17.63$ \\
            & \naivemeanmodel & $28.08$ & $4.81$ & $16.77$ && $\mathbf{34.0}$ & $\underline{6.30}$ & $\underline{18.81}$ && $\underline{30.97}$ & $5.25$ & $\underline{18.36}$ \\
            & \syntheticmodel & $29.39$ & $4.68$ & $17.35$ && $33.62$ & $6.06$ & $18.61$ && $30.65$ & $5.23$ & $18.11$ \\
            & \inductivebiasmodel & $28.41$ & $4.65$ & $16.90$ && $\underline{33.95}$ & $\mathbf{6.40}$ & $\mathbf{19.23}$ && $\mathbf{31.89}$ & $\mathbf{5.78}$ & $\mathbf{18.84}$ \\
            \bottomrule
        \end{tabular}
    }
    \caption{Reference-based Evaluation Results (R-$1$: \rouge{1}, R-$2$: \rouge{2}, R-L: \rouge{L}) for the \amazonb, \amazonbr \:and \amazonbrdq \:benchmarks. We see the following things: (\textit{a}) Our proposed dataset (\propdata) leads to \textit{marked increased over the SOTA} (by $\sim4$ R-L points), (\textit{b}) \inductivebiasmodel proves to be the \textit{winner in all the four scenarios}: \scenei, \sceneii, \sceneiii \:and \sceneiv \:(Section \ref{subsec:models-list}), \textit{proving the efficacy of our technique}. We also see that for the \amazonb \:benchmark, our models lag behind. However, \textit{this is expected}, as we highlight in Section \ref{subsec:benchmark-datasets}.}
    \label{tab:automatic-eval-amazon}
\end{table*}

\subsection{Evaluation Results}\label{subsec:eval-results}

We test our approach on $\mathbf{9}$ \textbf{benchmarks} (Section \ref{subsec:benchmark-datasets}). In the main manuscript, we report automatic evaluation results on Amazon-based benchmarks (Table \ref{tab:automatic-eval-amazon}), human evaluation on the \amazonb \:benchmark, and \gpt{4} evaluations on \amazonb, \flipkartb \:and \oposumb \:benchmarks. \citet{liu-etal-2023-geval} show that GPT-$4$ evaluations correlate well with human evaluations for summarization; hence, in the interest of time and monetary expense, we resort to \gpt{4} evaluation. We include automatic evaluation results for the rest of the benchmarks (Tables \ref{tab:automatic-eval-flipkart} and \ref{tab:automatic-eval-oposum}), and \textsc{BertScore} based evaluations (Table \ref{tab:auto-eval-bertscore}) for all the benchmarks in Appendix \ref{asec:auto-eval-fo-benchmarks}. We also include model generations for a randomly sampled product in Appendix \ref{asec:auto-eval-fo-benchmarks} (Table \ref{tab:gen-example}). Due to the shortcomings highlighted in Section \ref{subsec:benchmark-datasets} and Appendix \ref{asec:benchmark-problems}, we complement our automatic evaluations of \amazonb, \flipkartb \:and \oposumb \:with human and \gpt{4} evaluations.

\noindent\textbf{\textit{\underline{Automatic Evaluation}}.}\: From Table \ref{tab:automatic-eval-amazon}, we see that our proposed models are always better than the SOTA for \amazonbr \:and \amazonbrdq. Supervised Fine Tuning on \propdata \:(\supmodel \:model) helps achieve significantly better \textsc{Rouge} scores. This highlights the efficacy of our proposed \propdata \:dataset. From the automatic evaluations on \amazonbr \:and \amazonbrdq, we see the following things: 

\noindent\textbf{Answer to \scenei}: We see that \inductivebiasmodel \:achieves gains over \supmodel. This answers the question in \scenei: Our technique is effective over and above using a good dataset.

\noindent\textbf{Answer to \sceneii}: We see that \inductivebiasmodel \:achieves gains over \naivemeanmodel. This answers the question in \sceneii: Our technique is effective over vanilla RL.

\noindent\textbf{Answer to \sceneiii}: We see that \inductivebiasmodel \:achieves gains over \syntheticmodel. This answers the question in \sceneiii: Our technique is effective over the SOTA RLHF technique, which works without human preference data.

\noindent\textbf{Answer to \sceneiv}: We see that \inductivebiasmodel \:achieves gains over \dpomodel. This verifies that gains of \inductivebiasmodel \:can be safely attributed to the approach (not just the quality of \opinpref).


\noindent\textbf{\textit{\underline{Human/\gpt{4} Evaluation}}.}\: We conduct human evaluation (Figure \ref{fig:human-eval-bar-chart}) for the \amazonb \:benchmark, using $3$ domain experts (details in Appendix \ref{asec:annotator-details}). We observe a Fleiss' Kappa ($\kappa$) score of $56.25\%$ (moderate agreement; agreement is moderate when $40\% \leq \kappa < 60\%$). We ask the experts to rank the summaries (anonymized and shuffled) given the reviews. Given the rankings, we compute the fraction of pairwise wins, ties, and losses among all the models. We compare summaries from \supmodel, \naivemeanmodel, \syntheticmodel, \inductivebiasmodel, \tjmodel \:(SOTA) models and ground truth summaries. We include ground truth summaries in the evaluation to verify our claims about the quality of the benchmarks. From Figure \ref{fig:human-eval-bar-chart}, we see that \inductivebiasmodel \:wins significantly over the competitors, further proving the efficacy of our technique.

\begin{figure}[h!]
    \centering
    \includegraphics[width=\linewidth]{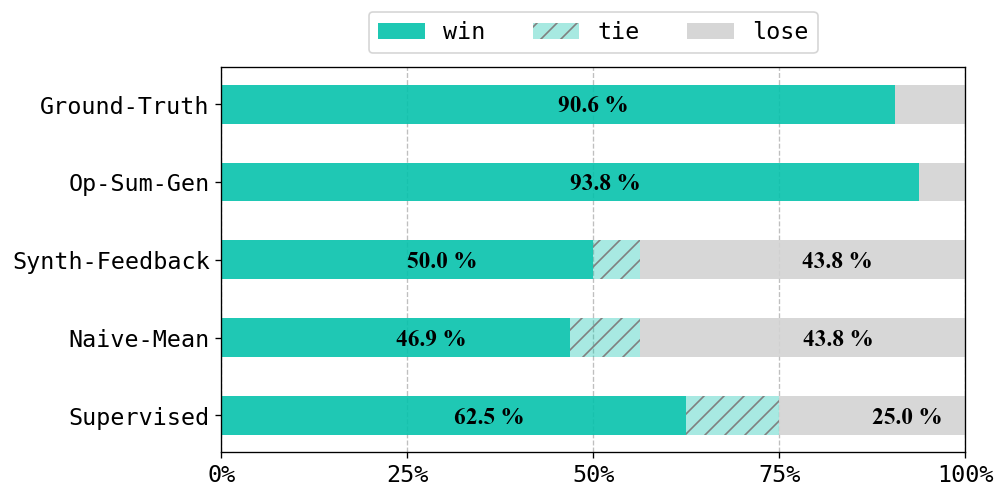}
    \caption{\gpt{4} Eval: Pairwise win-tie-loss percentage of \inductivebiasmodel \:model vs. competitors, for \amazonb \:benchmark.}
    \label{fig:gpt4-eval-amazon-eval-bar-chart}
\end{figure}

\begin{figure}[h!]
    \centering
    \includegraphics[width=\linewidth]{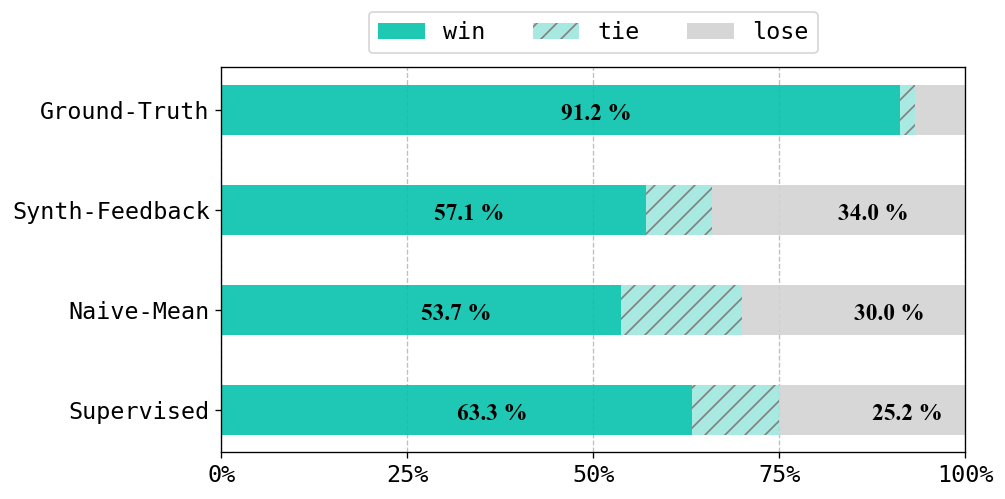}
    \caption{\gpt{4} Eval: Pairwise win-tie-loss percentage of \inductivebiasmodel \:model vs. competitors, for \flipkartb \:benchmark. Note that for the \flipkartb \:benchmark, we do not have results from \tjmodel, as \citet{siledar-etal-2023-synthesize} only provide aspect-specific summaries.}
    \label{fig:gpt4-eval-flipkart-bar-chart}
\end{figure}

\begin{figure}[h!]
    \centering
    \includegraphics[width=\linewidth]{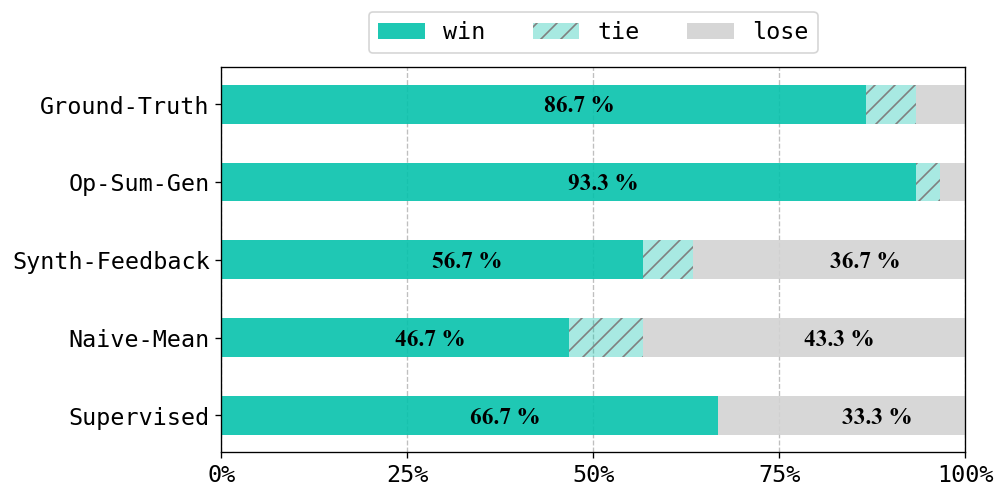}
    \caption{\gpt{4} Eval: Pairwise win-tie-loss percentage of \inductivebiasmodel \:model vs. competitors, for \oposumb \:benchmark.}
    \label{fig:gpt4-eval-oposum-bar-chart}
\end{figure}

We run \gpt{4} evaluations for \amazonb, \flipkartb \:and \oposumb \:benchmarks (Figures \ref{fig:gpt4-eval-amazon-eval-bar-chart}, \ref{fig:gpt4-eval-flipkart-bar-chart}, \ref{fig:gpt4-eval-oposum-bar-chart}). We run \gpt{4} evaluations for \amazonb, as the agreement in human evaluation was moderate. We arrive at the same conclusions as human evaluation. We prompt \gpt{4} to rank the summaries (anonymized and shuffled) given the reviews. As before, we compute the fraction of wins, ties, and losses. Again, we see that \inductivebiasmodel \:remains a clear winner.

\section{Analysis}\label{sec:analysis}

We perform a two-fold analysis: (\textit{a}) First, we see the domain knowledge features influence for \rmop, (\textit{b}) Second, we see how the ground truth summary and summary from trained models fare on the domain knowledge features. This two-fold analysis helps us understand: (\textit{a}) which features influence the latent reward model within humans\footnote{Note that the trained \rmop \:represents latent human reward model.} the most, and (\textit{b}) how the ground truth summary and summary from trained models fare on these influential features. Performing well on influential features would mean the summary aligns well with the latent reward model within humans.
\subsection{Analysis of \rmop}
\rmop \:model has been trained on a set of features specified by domain experts. We analyze the relative influence of each feature on the final score assigned by \rmop. Doing this helps us understand an approximate importance\footnote{We call this approximate importance as the influence of a feature on the output is not necessarily its importance.} of each of these features. We do this by varying each feature by $\delta \> (= 0.1)$ while keeping the other features constant, over multiple possible values of all features (Equation \ref{eqn:feature-influence}).

\vspace{-1em}
\begin{align}\label{eqn:feature-influence}
    \Delta_i = \dfrac{1}{2\delta}&\sum_{\mathbf{x}}\big(f(x_1, \cdots, x_i + \delta, \cdots, x_n)  \nonumber \\
    &- f(x_1, \cdots, x_i - \delta, \cdots, x_n)\big)
\end{align}

\begin{figure}[h]
    \centering
    \includegraphics[width=\linewidth]{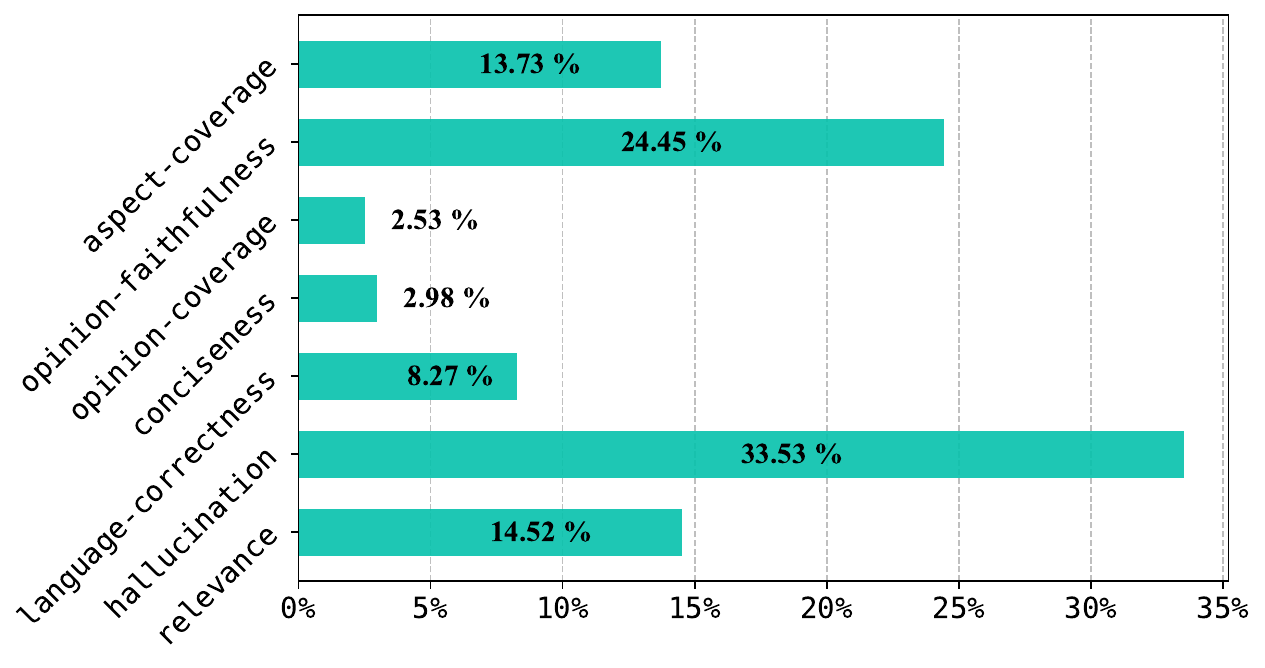}
    \caption{Relative Influence of all features in \rmop. All the influences sum to $1$.}
    \label{fig:feature-influence-reward-model}
\end{figure}

Figure \ref{fig:feature-influence-reward-model} highlights the features' relative influence. We see that \mhallucination \:is most influential. This aligns with what our human preference annotators report---hallucination in summary is the primary cause of rejection. We see that hallucinations are mostly within the opinions in the summary. This is also reflected in Figure \ref{fig:feature-influence-reward-model}: \mopfaith \:has significant influence. We also see that annotators prefer summaries with more specifics, i.e. they include more aspects: \maspcov \:has significant influence.

\subsection{Analysis of Summaries}

We analyze the top-$3$ performing models (in human and \gpt{4} evaluations) for the following features: \mopcov, \mopfaith, \mhallucination \:and \mrelevance. We show the analysis only for the \amazonb \:benchmark in the main manuscript, we include the rest in Appendix \ref{asec:all-eval-results}. Table \ref{tab:instrinsic-eval-plot-main-paper} shows the performance on these features. We see that \inductivebiasmodel \:model fares much better than the competitors on \mhallucination \:(the most influential metric). For \mrelevance, \maspcov \:and \mopfaith, our model is fairly better than the other models.

\begin{table}[h]
    \centering
    \begin{tabular}{*5c}
        \toprule
        Models & AC $\uparrow$ & OPF $\uparrow$ & RE $\uparrow$ & HL $\uparrow$ \\
        \midrule
        \texttt{IB} & $\mathbf{3.60}$ & $\mathbf{3.93}$ & $\mathbf{4.06}$ & $\mathbf{4.07}$\\
        \texttt{SF} & $3.43$ & $3.73$ & $4.04$ & $3.94$\\
        \texttt{NM} & $3.57$ & $3.91$ & $4.04$ & $3.09$\\
        \bottomrule
    \end{tabular}
    \caption{Scores on domain knowledge-based features (AC: \maspcov, RE: \mrelevance, OPF: \mopfaith, HL: \mhallucination) on the \amazonb \:benchmark for top-$3$ models ({\tt IB}: \inductivebiasmodel, {\tt NM}: \naivemeanmodel, {\tt SF}: \syntheticmodel). Note that for \mhallucination, $\Phi$ gives a higher score for less hallucination in the text.}
    \label{tab:instrinsic-eval-plot-main-paper}
\end{table}

This shows that our technique helps \inductivebiasmodel \:model perform well on features that influence the latent reward model within humans for opinion summarization. This means that our technique helps \inductivebiasmodel \:model achieve a significant alignment with the latent reward model. This conclusion \textbf{verifies our hypothesis} (in the domain of opinion summarization): \textit{A domain-knowledge infused reward model (\rmop) \underline{can help achieve alignment} with latent reward model of humans for a task, with \underline{modest} human preference annotations}.

\section{Summary, Conclusion and Future Work}\label{sec:summ-conc-future}
In this work, we propose a novel Reward Modeling technique via Domain Knowledge Infusion. We verify our approach for E-Commerce Opinion Summarization, where we achieve State-of-the-Art, while significantly reducing the amount of human preference annotations required (just $940$ samples). In addition to advancing SOTA and reducing preference annotations, our technique provides another two-fold benefits: (\textit{a}) No Alignment Tax and (\textit{b}) Interpretability. Due to the interpretable nature, we find that our model does achieve alignment with human goals for Opinion Summarization through analysis. From the results and analysis, we conclude that Domain Knowledge Infusion into Reward Modeling is a viable solution to reduce human preference annotations for downstream tasks. In the future, we will verify this for other domains.

\section{Ethical Considerations}
We contribute two datasets in our work: \propdata, \opinpref. These datasets are generated using an open-source model \mistral{7} \cite{jiang-etal-2023-mistral}. We would release the datasets to further research in Opinion Summarization. For the \opinpref, to the best of our knowledge, we have seen that it does not contain any harmful content, such as social biases, stereotypes, etc. However, we have seen that it contains products of explicit nature (sexual products). For the \propdata \:dataset, to the best of our knowledge, there is no presence of harmful content, such as social biases, stereotypes etc. We urge the research community to use the datasets with caution and check for potential harmfulness, based on their use-cases.

\section{Limitations}
A limitation of our work is we have tested our approach for one domain: Opinion Summarization. However, we do not believe that this weakens our argument, as we have exhaustively shown that our approach not only advances SOTA but also interpretably achieves alignment with humans. Future work in other domains would help in verifying this claim for other domains. Another limitation is: we see empirically that $\Phi$ works well for Opinion Summarization, to extract the scores for the $7$ features. However, there is no guarantee that such out-of-the-box performance would be reflected in another domain. Some fine-tuning might be necessary.

\bibliography{custom, anthology}

\begin{thebibliography}{39}
\expandafter\ifx\csname natexlab\endcsname\relax\def\natexlab#1{#1}\fi

\bibitem[{Amplayo et~al.(2021)Amplayo, Angelidis, and Lapata}]{amplayo-etal-2021-aspect}
Reinald~Kim Amplayo, Stefanos Angelidis, and Mirella Lapata. 2021.
\newblock \href {https://doi.org/10.18653/v1/2021.emnlp-main.528} {Aspect-controllable opinion summarization}.
\newblock In \emph{Proceedings of the 2021 Conference on Empirical Methods in Natural Language Processing}, pages 6578--6593, Online and Punta Cana, Dominican Republic. Association for Computational Linguistics.

\bibitem[{Amplayo and Lapata(2020)}]{amplayo-lapata-2020-unsupervised}
Reinald~Kim Amplayo and Mirella Lapata. 2020.
\newblock \href {https://doi.org/10.18653/v1/2020.acl-main.175} {Unsupervised opinion summarization with noising and denoising}.
\newblock In \emph{Proceedings of the 58th Annual Meeting of the Association for Computational Linguistics}, pages 1934--1945, Online. Association for Computational Linguistics.

\bibitem[{Askell et~al.(2021)Askell, Bai, Chen, Drain, Ganguli, Henighan, Jones, Joseph, Mann, DasSarma, Elhage, Hatfield-Dodds, Hernandez, Kernion, Ndousse, Olsson, Amodei, Brown, Clark, McCandlish, Olah, and Kaplan}]{askell-etal-2021-general}
Amanda Askell, Yuntao Bai, Anna Chen, Dawn Drain, Deep Ganguli, Tom Henighan, Andy Jones, Nicholas Joseph, Ben Mann, Nova DasSarma, Nelson Elhage, Zac Hatfield-Dodds, Danny Hernandez, Jackson Kernion, Kamal Ndousse, Catherine Olsson, Dario Amodei, Tom Brown, Jack Clark, Sam McCandlish, Chris Olah, and Jared Kaplan. 2021.
\newblock \href {http://arxiv.org/abs/2112.00861} {A general language assistant as a laboratory for alignment}.

\bibitem[{Bai et~al.(2022{\natexlab{a}})Bai, Jones, Ndousse, Askell, Chen, DasSarma, Drain, Fort, Ganguli, Henighan, Joseph, Kadavath, Kernion, Conerly, El-Showk, Elhage, Hatfield-Dodds, Hernandez, Hume, Johnston, Kravec, Lovitt, Nanda, Olsson, Amodei, Brown, Clark, McCandlish, Olah, Mann, and Kaplan}]{bai-etal-2022-hh-rlhf-data}
Yuntao Bai, Andy Jones, Kamal Ndousse, Amanda Askell, Anna Chen, Nova DasSarma, Dawn Drain, Stanislav Fort, Deep Ganguli, T.~J. Henighan, Nicholas Joseph, Saurav Kadavath, John Kernion, Tom Conerly, Sheer El-Showk, Nelson Elhage, Zac Hatfield-Dodds, Danny Hernandez, Tristan Hume, Scott Johnston, Shauna Kravec, Liane Lovitt, Neel Nanda, Catherine Olsson, Dario Amodei, Tom~B. Brown, Jack Clark, Sam McCandlish, Christopher Olah, Benjamin Mann, and Jared Kaplan. 2022{\natexlab{a}}.
\newblock \href {https://api.semanticscholar.org/CorpusID:248118878} {Training a helpful and harmless assistant with reinforcement learning from human feedback}.
\newblock \emph{ArXiv}, abs/2204.05862.

\bibitem[{Bai et~al.(2022{\natexlab{b}})Bai, Kadavath, Kundu, Askell, Kernion, Jones, Chen, Goldie, Mirhoseini, McKinnon, Chen, Olsson, Olah, Hernandez, Drain, Ganguli, Li, Tran-Johnson, Perez, Kerr, Mueller, Ladish, Landau, Ndousse, Lukosuite, Lovitt, Sellitto, Elhage, Schiefer, Mercado, DasSarma, Lasenby, Larson, Ringer, Johnston, Kravec, Showk, Fort, Lanham, Telleen-Lawton, Conerly, Henighan, Hume, Bowman, Hatfield-Dodds, Mann, Amodei, Joseph, McCandlish, Brown, and Kaplan}]{bai-etal-2022-constitutional-ai}
Yuntao Bai, Saurav Kadavath, Sandipan Kundu, Amanda Askell, Jackson Kernion, Andy Jones, Anna Chen, Anna Goldie, Azalia Mirhoseini, Cameron McKinnon, Carol Chen, Catherine Olsson, Christopher Olah, Danny Hernandez, Dawn Drain, Deep Ganguli, Dustin Li, Eli Tran-Johnson, Ethan Perez, Jamie Kerr, Jared Mueller, Jeffrey Ladish, Joshua Landau, Kamal Ndousse, Kamile Lukosuite, Liane Lovitt, Michael Sellitto, Nelson Elhage, Nicholas Schiefer, Noemi Mercado, Nova DasSarma, Robert Lasenby, Robin Larson, Sam Ringer, Scott Johnston, Shauna Kravec, Sheer~El Showk, Stanislav Fort, Tamera Lanham, Timothy Telleen-Lawton, Tom Conerly, Tom Henighan, Tristan Hume, Samuel~R. Bowman, Zac Hatfield-Dodds, Ben Mann, Dario Amodei, Nicholas Joseph, Sam McCandlish, Tom Brown, and Jared Kaplan. 2022{\natexlab{b}}.
\newblock \href {http://arxiv.org/abs/2212.08073} {Constitutional ai: Harmlessness from ai feedback}.

\bibitem[{Bhaskar et~al.(2023)Bhaskar, Fabbri, and Durrett}]{bhaskar-etal-2023-prompted}
Adithya Bhaskar, Alex Fabbri, and Greg Durrett. 2023.
\newblock \href {https://doi.org/10.18653/v1/2023.findings-acl.591} {Prompted opinion summarization with {GPT}-3.5}.
\newblock In \emph{Findings of the Association for Computational Linguistics: ACL 2023}, pages 9282--9300, Toronto, Canada. Association for Computational Linguistics.

\bibitem[{Bradley and Terry(1952)}]{bradley-terry-human-pref-model}
Ralph~Allan Bradley and Milton~E. Terry. 1952.
\newblock \href {http://www.jstor.org/stable/2334029} {Rank analysis of incomplete block designs: I. the method of paired comparisons}.
\newblock \emph{Biometrika}, 39(3/4):324--345.

\bibitem[{Bra{\v{z}}inskas et~al.(2020)Bra{\v{z}}inskas, Lapata, and Titov}]{brazinskas-etal-2020-unsupervised}
Arthur Bra{\v{z}}inskas, Mirella Lapata, and Ivan Titov. 2020.
\newblock \href {https://doi.org/10.18653/v1/2020.acl-main.461} {Unsupervised opinion summarization as copycat-review generation}.
\newblock In \emph{Proceedings of the 58th Annual Meeting of the Association for Computational Linguistics}, pages 5151--5169, Online. Association for Computational Linguistics.

\bibitem[{Chiang et~al.(2023)Chiang, Li, Lin, Sheng, Wu, Zhang, Zheng, Zhuang, Zhuang, Gonzalez, Stoica, and Xing}]{chiang-etal-2023-vicuna}
Wei-Lin Chiang, Zhuohan Li, Zi~Lin, Ying Sheng, Zhanghao Wu, Hao Zhang, Lianmin Zheng, Siyuan Zhuang, Yonghao Zhuang, Joseph~E. Gonzalez, Ion Stoica, and Eric~P. Xing. 2023.
\newblock \href {https://lmsys.org/blog/2023-03-30-vicuna/} {Vicuna: An open-source chatbot impressing gpt-4 with 90\%* chatgpt quality}.

\bibitem[{Chu and Liu(2018)}]{chu-etal-2019-mean-sum}
Eric Chu and Peter~J. Liu. 2018.
\newblock \href {https://api.semanticscholar.org/CorpusID:59413781} {Meansum: A neural model for unsupervised multi-document abstractive summarization}.
\newblock In \emph{International Conference on Machine Learning}.

\bibitem[{Ethayarajh et~al.(2022)Ethayarajh, Choi, and Swayamdipta}]{chaudhari-etal-2022-shp-data}
Kawin Ethayarajh, Yejin Choi, and Swabha Swayamdipta. 2022.
\newblock \href {https://proceedings.mlr.press/v162/ethayarajh22a.html} {Understanding dataset difficulty with $\mathcal{V}$-usable information}.
\newblock In \emph{Proceedings of the 39th International Conference on Machine Learning}, volume 162 of \emph{Proceedings of Machine Learning Research}, pages 5988--6008. PMLR.

\bibitem[{Glaese et~al.(2022)Glaese, McAleese, Trębacz, Aslanides, Firoiu, Ewalds, Rauh, Weidinger, Chadwick, Thacker, Campbell-Gillingham, Uesato, Huang, Comanescu, Yang, See, Dathathri, Greig, Chen, Fritz, Elias, Green, Mokrá, Fernando, Wu, Foley, Young, Gabriel, Isaac, Mellor, Hassabis, Kavukcuoglu, Hendricks, and Irving}]{glaese-etal-2022-sparrow}
Amelia Glaese, Nat McAleese, Maja Trębacz, John Aslanides, Vlad Firoiu, Timo Ewalds, Maribeth Rauh, Laura Weidinger, Martin Chadwick, Phoebe Thacker, Lucy Campbell-Gillingham, Jonathan Uesato, Po-Sen Huang, Ramona Comanescu, Fan Yang, Abigail See, Sumanth Dathathri, Rory Greig, Charlie Chen, Doug Fritz, Jaume~Sanchez Elias, Richard Green, Soňa Mokrá, Nicholas Fernando, Boxi Wu, Rachel Foley, Susannah Young, Iason Gabriel, William Isaac, John Mellor, Demis Hassabis, Koray Kavukcuoglu, Lisa~Anne Hendricks, and Geoffrey Irving. 2022.
\newblock \href {http://arxiv.org/abs/2209.14375} {Improving alignment of dialogue agents via targeted human judgements}.

\bibitem[{He and McAuley(2016)}]{he-mcaulay-2016-amazon}
Ruining He and Julian McAuley. 2016.
\newblock \href {https://doi.org/10.1145/2872427.2883037} {Ups and downs: Modeling the visual evolution of fashion trends with one-class collaborative filtering}.
\newblock In \emph{Proceedings of the 25th International Conference on World Wide Web}, WWW '16, page 507–517, Republic and Canton of Geneva, CHE. International World Wide Web Conferences Steering Committee.

\bibitem[{Hu and Liu(2004)}]{hu-liu-2004-opinion-mining}
Minqing Hu and Bing Liu. 2004.
\newblock \href {https://doi.org/10.1145/1014052.1014073} {Mining and summarizing customer reviews}.
\newblock In \emph{Proceedings of the Tenth ACM SIGKDD International Conference on Knowledge Discovery and Data Mining}, KDD '04, page 168–177, New York, NY, USA. Association for Computing Machinery.

\bibitem[{Im et~al.(2021)Im, Kim, Lee, Cho, and Chung}]{im-etal-2021-self}
Jinbae Im, Moonki Kim, Hoyeop Lee, Hyunsouk Cho, and Sehee Chung. 2021.
\newblock \href {https://doi.org/10.18653/v1/2021.acl-long.33} {Self-supervised multimodal opinion summarization}.
\newblock In \emph{Proceedings of the 59th Annual Meeting of the Association for Computational Linguistics and the 11th International Joint Conference on Natural Language Processing (Volume 1: Long Papers)}, pages 388--403, Online. Association for Computational Linguistics.

\bibitem[{Jiang et~al.(2023)Jiang, Sablayrolles, Mensch, Bamford, Chaplot, de~las Casas, Bressand, Lengyel, Lample, Saulnier, Lavaud, Lachaux, Stock, Scao, Lavril, Wang, Lacroix, and Sayed}]{jiang-etal-2023-mistral}
Albert~Q. Jiang, Alexandre Sablayrolles, Arthur Mensch, Chris Bamford, Devendra~Singh Chaplot, Diego de~las Casas, Florian Bressand, Gianna Lengyel, Guillaume Lample, Lucile Saulnier, Lélio~Renard Lavaud, Marie-Anne Lachaux, Pierre Stock, Teven~Le Scao, Thibaut Lavril, Thomas Wang, Timothée Lacroix, and William~El Sayed. 2023.
\newblock \href {http://arxiv.org/abs/2310.06825} {Mistral 7b}.

\bibitem[{Jiang et~al.(2022)Jiang, Hwang, Bhagavatula, Bras, Liang, Dodge, Sakaguchi, Forbes, Borchardt, Gabriel, Tsvetkov, Etzioni, Sap, Rini, and Choi}]{jiang-etal-2022-delphi}
Liwei Jiang, Jena~D. Hwang, Chandra Bhagavatula, Ronan~Le Bras, Jenny Liang, Jesse Dodge, Keisuke Sakaguchi, Maxwell Forbes, Jon Borchardt, Saadia Gabriel, Yulia Tsvetkov, Oren Etzioni, Maarten Sap, Regina Rini, and Yejin Choi. 2022.
\newblock \href {http://arxiv.org/abs/2110.07574} {Can machines learn morality? the delphi experiment}.

\bibitem[{Kim et~al.(2023)Kim, Bae, Shin, Kang, Kwak, Yoo, and Seo}]{kim-etal-2023-aligning}
Sungdong Kim, Sanghwan Bae, Jamin Shin, Soyoung Kang, Donghyun Kwak, Kang Yoo, and Minjoon Seo. 2023.
\newblock \href {https://doi.org/10.18653/v1/2023.emnlp-main.844} {Aligning large language models through synthetic feedback}.
\newblock In \emph{Proceedings of the 2023 Conference on Empirical Methods in Natural Language Processing}, pages 13677--13700, Singapore. Association for Computational Linguistics.

\bibitem[{Lee et~al.(2023)Lee, Phatale, Mansoor, Mesnard, Ferret, Lu, Bishop, Hall, Carbune, Rastogi, and Prakash}]{lee-etal-2023-rlaif}
Harrison Lee, Samrat Phatale, Hassan Mansoor, Thomas Mesnard, Johan Ferret, Kellie Lu, Colton Bishop, Ethan Hall, Victor Carbune, Abhinav Rastogi, and Sushant Prakash. 2023.
\newblock \href {http://arxiv.org/abs/2309.00267} {Rlaif: Scaling reinforcement learning from human feedback with ai feedback}.

\bibitem[{Lewis et~al.(2020)Lewis, Liu, Goyal, Ghazvininejad, Mohamed, Levy, Stoyanov, and Zettlemoyer}]{lewis-etal-2020-bart}
Mike Lewis, Yinhan Liu, Naman Goyal, Marjan Ghazvininejad, Abdelrahman Mohamed, Omer Levy, Veselin Stoyanov, and Luke Zettlemoyer. 2020.
\newblock \href {https://doi.org/10.18653/v1/2020.acl-main.703} {{BART}: Denoising sequence-to-sequence pre-training for natural language generation, translation, and comprehension}.
\newblock In \emph{Proceedings of the 58th Annual Meeting of the Association for Computational Linguistics}, pages 7871--7880, Online. Association for Computational Linguistics.

\bibitem[{Liu et~al.(2022)Liu, Zhang, Feng, and Vosoughi}]{liu-etal-2022-aligning}
Ruibo Liu, Ge~Zhang, Xinyu Feng, and Soroush Vosoughi. 2022.
\newblock \href {https://doi.org/10.18653/v1/2022.findings-naacl.18} {Aligning generative language models with human values}.
\newblock In \emph{Findings of the Association for Computational Linguistics: NAACL 2022}, pages 241--252, Seattle, United States. Association for Computational Linguistics.

\bibitem[{Liu et~al.(2023)Liu, Iter, Xu, Wang, Xu, and Zhu}]{liu-etal-2023-geval}
Yang Liu, Dan Iter, Yichong Xu, Shuohang Wang, Ruochen Xu, and Chenguang Zhu. 2023.
\newblock \href {https://doi.org/10.18653/v1/2023.emnlp-main.153} {{G}-eval: {NLG} evaluation using gpt-4 with better human alignment}.
\newblock In \emph{Proceedings of the 2023 Conference on Empirical Methods in Natural Language Processing}, pages 2511--2522, Singapore. Association for Computational Linguistics.

\bibitem[{Luce(2012)}]{luce-human-pref-model}
R.D. Luce. 2012.
\newblock \href {https://books.google.co.in/books?id=ERQsKkPiKkkC} {\emph{Individual Choice Behavior: A Theoretical Analysis}}.
\newblock Dover Books on Mathematics. Dover Publications.

\bibitem[{Nakano et~al.(2021)Nakano, Hilton, Balaji, Wu, Long, Kim, Hesse, Jain, Kosaraju, Saunders, Jiang, Cobbe, Eloundou, Krueger, Button, Knight, Chess, and Schulman}]{nakano-etal-2021-webgpt}
Reiichiro Nakano, Jacob Hilton, Suchir Balaji, Jeff Wu, Ouyang Long, Christina Kim, Christopher Hesse, Shantanu Jain, Vineet Kosaraju, William Saunders, Xu~Jiang, Karl Cobbe, Tyna Eloundou, Gretchen Krueger, Kevin Button, Matthew Knight, Benjamin Chess, and John Schulman. 2021.
\newblock \href {https://api.semanticscholar.org/CorpusID:245329531} {Webgpt: Browser-assisted question-answering with human feedback}.
\newblock \emph{ArXiv}, abs/2112.09332.

\bibitem[{Ouyang et~al.(2022)Ouyang, Wu, Jiang, Almeida, Wainwright, Mishkin, Zhang, Agarwal, Slama, Ray, Schulman, Hilton, Kelton, Miller, Simens, Askell, Welinder, Christiano, Leike, and Lowe}]{ouyang-etal-2022-instruct-gpt}
Long Ouyang, Jeffrey Wu, Xu~Jiang, Diogo Almeida, Carroll Wainwright, Pamela Mishkin, Chong Zhang, Sandhini Agarwal, Katarina Slama, Alex Ray, John Schulman, Jacob Hilton, Fraser Kelton, Luke Miller, Maddie Simens, Amanda Askell, Peter Welinder, Paul~F Christiano, Jan Leike, and Ryan Lowe. 2022.
\newblock \href {https://proceedings.neurips.cc/paper_files/paper/2022/file/b1efde53be364a73914f58805a001731-Paper-Conference.pdf} {Training language models to follow instructions with human feedback}.
\newblock In \emph{Advances in Neural Information Processing Systems}, volume~35, pages 27730--27744. Curran Associates, Inc.

\bibitem[{Peng et~al.(2023)Peng, Li, He, Galley, and Gao}]{peng-etal-2023-instruction}
Baolin Peng, Chunyuan Li, Pengcheng He, Michel Galley, and Jianfeng Gao. 2023.
\newblock Instruction tuning with gpt-4.
\newblock \emph{arXiv preprint arXiv:2304.03277}.

\bibitem[{Plackett(1975)}]{plackett-human-pref-model}
R.~L. Plackett. 1975.
\newblock \href {http://www.jstor.org/stable/2346567} {The analysis of permutations}.
\newblock \emph{Journal of the Royal Statistical Society. Series C (Applied Statistics)}, 24(2):193--202.

\bibitem[{Rafailov et~al.(2023)Rafailov, Sharma, Mitchell, Ermon, Manning, and Finn}]{rafailov-etal-2023-dpo}
Rafael Rafailov, Archit Sharma, Eric Mitchell, Stefano Ermon, Christopher~D. Manning, and Chelsea Finn. 2023.
\newblock \href {http://arxiv.org/abs/2305.18290} {Direct preference optimization: Your language model is secretly a reward model}.

\bibitem[{Schulman et~al.(2017)Schulman, Wolski, Dhariwal, Radford, and Klimov}]{schulman-etal-2017-ppo}
John Schulman, Filip Wolski, Prafulla Dhariwal, Alec Radford, and Oleg Klimov. 2017.
\newblock \href {http://arxiv.org/abs/1707.06347} {Proximal policy optimization algorithms}.

\bibitem[{Siledar et~al.(2023{\natexlab{a}})Siledar, Banerjee, Patil, Singh, Chelliah, Garera, and Bhattacharyya}]{siledar-etal-2023-synthesize}
Tejpalsingh Siledar, Suman Banerjee, Amey Patil, Sudhanshu Singh, Muthusamy Chelliah, Nikesh Garera, and Pushpak Bhattacharyya. 2023{\natexlab{a}}.
\newblock \href {https://doi.org/10.18653/v1/2023.findings-emnlp.899} {Synthesize, if you do not have: Effective synthetic dataset creation strategies for self-supervised opinion summarization in {E}-commerce}.
\newblock In \emph{Findings of the Association for Computational Linguistics: EMNLP 2023}, pages 13480--13491, Singapore. Association for Computational Linguistics.

\bibitem[{Siledar et~al.(2023{\natexlab{b}})Siledar, Makwana, and Bhattacharyya}]{siledar-etal-2023-opinion-summ}
Tejpalsingh Siledar, Jigar Makwana, and Pushpak Bhattacharyya. 2023{\natexlab{b}}.
\newblock \href {https://doi.org/10.1145/3570991.3571035} {Aspect-sentiment-based opinion summarization using multiple information sources}.
\newblock In \emph{Proceedings of the 6th Joint International Conference on Data Science {\&} Management of Data (10th {ACM} {IKDD} {CODS} and 28th COMAD), Mumbai, India, January 4-7, 2023}, pages 55--61. {ACM}.

\bibitem[{Siledar et~al.(2024)Siledar, Rangaraju, Muddu, Banerjee, Patil, Singh, Chelliah, Garera, Nath, and Bhattacharyya}]{siledar-etal-2024-product}
Tejpalsingh Siledar, Rupasai Rangaraju, Sankara Sri Raghava~Ravindra Muddu, Suman Banerjee, Amey Patil, Sudhanshu~Shekhar Singh, Muthusamy Chelliah, Nikesh Garera, Swaprava Nath, and Pushpak Bhattacharyya. 2024.
\newblock \href {http://arxiv.org/abs/2404.05243} {Product description and qa assisted self-supervised opinion summarization}.

\bibitem[{Sorensen et~al.(2023)Sorensen, Jiang, Hwang, Levine, Pyatkin, West, Dziri, Lu, Rao, Bhagavatula, Sap, Tasioulas, and Choi}]{sorensen-etal-2023-value-pluralism}
Taylor Sorensen, Liwei Jiang, Jena Hwang, Sydney Levine, Valentina Pyatkin, Peter West, Nouha Dziri, Ximing Lu, Kavel Rao, Chandra Bhagavatula, Maarten Sap, John Tasioulas, and Yejin Choi. 2023.
\newblock \href {http://arxiv.org/abs/2309.00779} {Value kaleidoscope: Engaging ai with pluralistic human values, rights, and duties}.

\bibitem[{Taori et~al.(2023)Taori, Gulrajani, Zhang, Dubois, Li, Guestrin, Liang, and Hashimoto}]{taori-etal-2023-alpaca}
Rohan Taori, Ishaan Gulrajani, Tianyi Zhang, Yann Dubois, Xuechen Li, Carlos Guestrin, Percy Liang, and Tatsunori~B. Hashimoto. 2023.
\newblock Stanford alpaca: An instruction-following llama model.
\newblock \url{https://github.com/tatsu-lab/stanford_alpaca}.

\bibitem[{Touvron et~al.(2023{\natexlab{a}})Touvron, Lavril, Izacard, Martinet, Lachaux, Lacroix, Rozière, Goyal, Hambro, Azhar, Rodriguez, Joulin, Grave, and Lample}]{touvron-etal-2023-llama}
Hugo Touvron, Thibaut Lavril, Gautier Izacard, Xavier Martinet, Marie-Anne Lachaux, Timothée Lacroix, Baptiste Rozière, Naman Goyal, Eric Hambro, Faisal Azhar, Aurelien Rodriguez, Armand Joulin, Edouard Grave, and Guillaume Lample. 2023{\natexlab{a}}.
\newblock \href {http://arxiv.org/abs/2302.13971} {Llama: Open and efficient foundation language models}.

\bibitem[{Touvron et~al.(2023{\natexlab{b}})Touvron, Martin, Stone, Albert, Almahairi, Babaei, Bashlykov, Batra, Bhargava, Bhosale, Bikel, Blecher, Ferrer, Chen, Cucurull, Esiobu, Fernandes, Fu, Fu, Fuller, Gao, Goswami, Goyal, Hartshorn, Hosseini, Hou, Inan, Kardas, Kerkez, Khabsa, Kloumann, Korenev, Koura, Lachaux, Lavril, Lee, Liskovich, Lu, Mao, Martinet, Mihaylov, Mishra, Molybog, Nie, Poulton, Reizenstein, Rungta, Saladi, Schelten, Silva, Smith, Subramanian, Tan, Tang, Taylor, Williams, Kuan, Xu, Yan, Zarov, Zhang, Fan, Kambadur, Narang, Rodriguez, Stojnic, Edunov, and Scialom}]{touvron-etal-2023-llama-two}
Hugo Touvron, Louis Martin, Kevin Stone, Peter Albert, Amjad Almahairi, Yasmine Babaei, Nikolay Bashlykov, Soumya Batra, Prajjwal Bhargava, Shruti Bhosale, Dan Bikel, Lukas Blecher, Cristian~Canton Ferrer, Moya Chen, Guillem Cucurull, David Esiobu, Jude Fernandes, Jeremy Fu, Wenyin Fu, Brian Fuller, Cynthia Gao, Vedanuj Goswami, Naman Goyal, Anthony Hartshorn, Saghar Hosseini, Rui Hou, Hakan Inan, Marcin Kardas, Viktor Kerkez, Madian Khabsa, Isabel Kloumann, Artem Korenev, Punit~Singh Koura, Marie-Anne Lachaux, Thibaut Lavril, Jenya Lee, Diana Liskovich, Yinghai Lu, Yuning Mao, Xavier Martinet, Todor Mihaylov, Pushkar Mishra, Igor Molybog, Yixin Nie, Andrew Poulton, Jeremy Reizenstein, Rashi Rungta, Kalyan Saladi, Alan Schelten, Ruan Silva, Eric~Michael Smith, Ranjan Subramanian, Xiaoqing~Ellen Tan, Binh Tang, Ross Taylor, Adina Williams, Jian~Xiang Kuan, Puxin Xu, Zheng Yan, Iliyan Zarov, Yuchen Zhang, Angela Fan, Melanie Kambadur, Sharan Narang, Aurelien Rodriguez, Robert Stojnic, Sergey Edunov, and Thomas
  Scialom. 2023{\natexlab{b}}.
\newblock \href {http://arxiv.org/abs/2307.09288} {Llama 2: Open foundation and fine-tuned chat models}.

\bibitem[{Tunstall et~al.(2023)Tunstall, Beeching, Lambert, Rajani, Rasul, Belkada, Huang, von Werra, Fourrier, Habib, Sarrazin, Sanseviero, Rush, and Wolf}]{tunstall-etal-2023-zephyr}
Lewis Tunstall, Edward Beeching, Nathan Lambert, Nazneen Rajani, Kashif Rasul, Younes Belkada, Shengyi Huang, Leandro von Werra, Clémentine Fourrier, Nathan Habib, Nathan Sarrazin, Omar Sanseviero, Alexander~M. Rush, and Thomas Wolf. 2023.
\newblock \href {http://arxiv.org/abs/2310.16944} {Zephyr: Direct distillation of lm alignment}.

\bibitem[{Wang et~al.(2023)Wang, Kordi, Mishra, Liu, Smith, Khashabi, and Hajishirzi}]{wang-etal-2023-self-instruct}
Yizhong Wang, Yeganeh Kordi, Swaroop Mishra, Alisa Liu, Noah~A. Smith, Daniel Khashabi, and Hannaneh Hajishirzi. 2023.
\newblock \href {https://doi.org/10.18653/v1/2023.acl-long.754} {Self-instruct: Aligning language models with self-generated instructions}.
\newblock In \emph{Proceedings of the 61st Annual Meeting of the Association for Computational Linguistics (Volume 1: Long Papers)}, pages 13484--13508, Toronto, Canada. Association for Computational Linguistics.

\bibitem[{Ziegler et~al.(2019)Ziegler, Stiennon, Wu, Brown, Radford, Amodei, Christiano, and Irving}]{ziegler-etal-2019-fine-tuning-lms-human-pref}
Daniel~M. Ziegler, Nisan Stiennon, Jeff Wu, Tom~B. Brown, Alec Radford, Dario Amodei, Paul Christiano, and Geoffrey Irving. 2019.
\newblock \href {https://api.semanticscholar.org/CorpusID:202660943} {Fine-tuning language models from human preferences}.
\newblock \emph{ArXiv}, abs/1909.08593.

\end{thebibliography}

\appendix

\section{Features for Reward Modeling}\label{asec:reward-modelling-featues}

We use $7$ domain specific features for the reward model \rmop. We identify these features after extensive discussions with the domain experts. For each feature we prompt \mistral{7} to generate a score within $0$ and $5$. We give elongated instructions, including rough rubriks, and rules to generate the scores. This is a reason why we use an instruction-tuned model. For each feature, $0$ means the model is doing bad on the feature, and $5$ means the model is doing good on the feature. We define all the features below:

\noindent\maspcovb: This feature considers the aspect coverage within an opinion summary. The feature assumes a value $5$ if all the aspects of the product, mentioned in the reviews, are mentioned in the summary. If none of the aspects are picked, the feature assumes a value $0$.

\noindent\mopfaithb: This feature considers whether the mentioned opinions/sentiments in the summary are correct, that is, they are picked correctly from the reviews. For example, if an user mentions that they are \textit{happy} with the battery of a phone, and the summary mentions that users are \textit{unhappy} with the battery, the summary will not be considered faithful to opinion in the review. The feature assumes a value $5$ if all the opinions are faithfully reflected. If no opinion is faithfully reflected, the value would be $0$.

\noindent\mopcovb: This feature considers whether all the opinions in the input reviews are picked by the opinion summary. The feature assumes a value $5$ if all the opinions are picked up. If none of the opinions are picked up, the feature assumes a value $0$.

\noindent\mrelevanceb: This feature checks if the summary is relevant to the input reviews (that is the product). The feature assumes a value $5$ if summary is completely relevant. If it is completely irrelevant, the feature assumes a value $0$.

\noindent\mconciseb: This feature considers the conciseness and completeness of the opinion summary. The feature assumes a value $5$ if the summary is concise and complete---not one phrase/sentence can be dropped off. It assumes a value $0$ if the summary is totally incomplete, or very verbose.

\noindent\mhallucinationb: This feature considers the factuality of the opinion summary. The feature assumes a value $5$ if the summary is totally factual, with respect to the input reviews. If there are a lot of hallucinations, the feature assumes a value $0$.

\noindent\mlangcorrb: This feature checks the correctness of language/text in the opinion summary. The feature assumes a value $5$ if the summary is grammatically fully correct. It assumes a value $0$ if the summary is very poor linguistically.

For conciseness, we do not include the prompts in the paper, we would release them as separate artifacts, with the datasets, in the camera ready version.

We also analyze how these features correlate with humans' judgement of goodness of opinion summaries. We do this by looking at the scores for these features for preferred and dis-preferred summaries in the \opinpref \:dataset. In Table \ref{tab:feature-correlation}, we see that the preferred summaries clearly have a higher score on all the features, than the dis-preferred ones. This shows that the scores correlate well with humans' judgement of goodness.

\begin{table}[h]
    \centering
    \begin{tabular}{*3c}
    \toprule
    Feature & Pref. & Dis-pref.\\
    \midrule
    \maspcov \:($\uparrow$) & $3.69$ & $2.84$ \\
    \mopfaith \:($\uparrow$) & $4.02$ & $3.05$ \\
    \mopcov \:($\uparrow$) & $3.92$ & $3.22$ \\
    \mconcise \:($\uparrow$) & $4.05$ & $3.44$ \\
    \mrelevance \:($\uparrow$) & $4.10$ & $3.10$ \\
    \mhallucination \:($\uparrow$) & $3.99$ & $2.79$ \\
    \mlangcorr \:($\uparrow$) & $4.50$ & $3.32$ \\
    \bottomrule
    \end{tabular}
    \caption{Scores for the domain knowledge based features. We see that for all the features, the human preferred (Pref.) summaries have higher scores than the ones rejected by humans (Dis-pref.). This shows that these features correlate well with humans' judgement of goodness of an opinion summary.}
    \label{tab:feature-correlation}
\end{table}

\begin{table*}[t]
    \centering
    \resizebox{2\columnwidth}{!}{
        \begin{tabular}{*{13}c}
            \toprule
            & \multirow{2}{*}{Model-Code} & \multicolumn{3}{c}{\flipkartb} && \multicolumn{3}{c}{\flipkartbr} && \multicolumn{3}{c}{\flipkartbrdq} \\
            \cmidrule{3-5}\cmidrule{7-9}\cmidrule{11-13}
            && R-$1 \uparrow$ & R-$2 \> \uparrow$ & R-L$\> \uparrow$ && R-$1 \> \uparrow$ & R-$2 \> \uparrow$ & R-L$\> \uparrow$ && R-$1 \> \uparrow$ & R-$2 \> \uparrow$ & R-L$\> \uparrow$ \\
            \midrule
            & \medosmodel \:\cite{siledar-etal-2024-product} & $25.97$ & $\mathbf{5.29}$ & $\underline{16.05}$ && $26.29$ & $4.03$ & $16.59$ && $22.92$ & $4.30$ & $16.35$ \\
            \midrule
            \multirow{5}{*}{\rot{\textit{Ours'}}}
            & \dpomodel & $28.85$ & $4.10$ & $15.55$ && $34.23$ & $7.86$ & $18.62$ && $29.96$ & $5.25$ & $17.28$ \\
            & \supmodel & $27.38$ & $4.09$ & $15.37$ && $39.32$ & $10.52$ & $22.56$ && $32.25$ & $6.88$ & $19.04$ \\
            & \naivemeanmodel & $28.34$ & $\underline{4.38}$ & $\mathbf{16.20}$ && $\mathbf{40.56}$ & $10.68$ & $22.74$ && $\underline{32.57}$ & $6.67$ & $\underline{19.39}$ \\
            & \syntheticmodel & $26.37$ & $4.18$ & $15.48$ && $38.77$ & $\underline{10.99}$ & $\underline{22.97}$ && $31.04$ & $\underline{6.98}$ & $18.59$ \\
            & \inductivebiasmodel & $\mathbf{27.42}$ & $4.21$ & $15.71$ && $\underline{39.10}$ & $\mathbf{11.03}$ & $\mathbf{23.30}$ && $\mathbf{33.08}$ & $\mathbf{7.30}$ & $\mathbf{19.46}$ \\
            \bottomrule
        \end{tabular}
    }
    \caption{Reference-based Evaluation Results (R-$1$: \rouge{1}, R-$2$: \rouge{2}, R-L: \rouge{L}) for the \flipkartb, \flipkartbr \:and \flipkartbrdq \:benchmarks. We see the following things: (\textit{a}) Our proposed dataset (\propdata) leads to \textit{marked increased over the SOTA} (\medosmodel; by $\sim6$ R-L points), (\textit{b}) \inductivebiasmodel proves to be the \textit{winner in all the four scenarios}: \scenei, \sceneii, \sceneiii \:and \sceneiv \:(Section \ref{subsec:models-list}), \textit{proving the efficacy of our technique}. We also see that for \flipkartb \:benchmark, despite the shortcomings, our models perform similar to the SOTA.}
    \label{tab:automatic-eval-flipkart}
\end{table*}

\begin{table*}[t]
    \centering
    \resizebox{2\columnwidth}{!}{
        \begin{tabular}{*{13}c}
            \toprule
            & \multirow{2}{*}{Model-Code} & \multicolumn{3}{c}{\oposumb} && \multicolumn{3}{c}{\oposumbr} && \multicolumn{3}{c}{\oposumbrdq} \\
            \cmidrule{3-5}\cmidrule{7-9}\cmidrule{11-13}
            && R-$1 \uparrow$ & R-$2 \> \uparrow$ & R-L$\> \uparrow$ && R-$1 \> \uparrow$ & R-$2 \> \uparrow$ & R-L$\> \uparrow$ && R-$1 \> \uparrow$ & R-$2 \> \uparrow$ & R-L$\> \uparrow$ \\
            \midrule
            \multirow{6}{*}{\rot{\textit{Prior Works}}} & MeanSum \cite{chu-etal-2019-mean-sum} & $26.25$ & $4.62$ & $16.49$ && $-$ & $-$ & $-$ && $-$ & $-$ & $-$ \\
            & CopyCat \cite{brazinskas-etal-2020-unsupervised} & $27.98$ & $5.79$ & $17.07$ && $22.41$ & $2.30$ & $13.94$ && $22.38$ & $2.03$ & $14.06$ \\
            & PlanSum \cite{amplayo-lapata-2020-unsupervised} & $30.26$ & $5.29$ & $17.48$ && $22.37$ & $2.05$ & $13.32$ && $22.64$ & $2.25$ & $13.71$ \\
            & MultimodalSum \cite{im-etal-2021-self} & $33.08$ & $7.46$ & $19.75$ && $23.35$ & $2.98$ & $14.53$ && $23.73$ & $2.80$ & $14.70$ \\
            & \tjmodel \:\cite{siledar-etal-2023-synthesize} & $36.44$ & $8.50$ & $22.03$ && $25.65$ & $3.56$ & $15.83$ && $24.66$ & $3.25$ & $15.54$ \\
            & \medosmodel \:\cite{siledar-etal-2024-product} & $\mathbf{36.57}$ & $\underline{8.79}$ & $\mathbf{21.35}$ && $26.82$ & $3.67$ & $15.92$ && $26.32$ & $3.34$ & $16.10$ \\
            \midrule
            \multirow{5}{*}{\rot{\textit{Ours'}}}
            & \dpomodel & $27.64$ & $7.34$ & $16.50$ && $33.69$ & $6.62$ & $18.55$ && $30.95$ & $5.89$ & $17.60$ \\
            & \supmodel & $30.57$ & $8.02$ & $16.90$ && $38.32$ & $9.10$ & $20.35$ && $35.69$ & $8.17$ & $19.28$ \\
            & \naivemeanmodel & $31.47$ & $8.0$ & $16.99$ && $40.16$ & $9.84$ & $21.74$ && $35.90$ & $8.33$ & $20.13$ \\
            & \syntheticmodel & $\underline{31.66}$ & $\mathbf{8.86}$ & $\underline{17.91}$ && $\underline{41.32}$ & $\mathbf{10.40}$ & $\mathbf{22.23}$ && $\mathbf{37.85}$ & $\underline{8.94}$ & $\underline{20.71}$ \\
            & \inductivebiasmodel & $31.15$ & $8.15$ & $17.46$ && $\mathbf{41.58}$ & $\underline{10.32}$ & $\underline{22.02}$ && $\underline{37.56}$ & $\mathbf{9.21}$ & $\mathbf{20.88}$ \\
            \bottomrule
        \end{tabular}
    }
    \caption{Reference-based Evaluation Results (R-$1$: \rouge{1}, R-$2$: \rouge{2}, R-L: \rouge{L}) for the \oposumb, \oposumbr \:and \oposumbrdq \:benchmarks. We see the following things: (\textit{a}) Our proposed dataset (\propdata) leads to \textit{marked increased over the SOTA} (\medosmodel; by $\sim6$ R-L points), (\textit{b}) \inductivebiasmodel proves to be the \textit{winner in \underline{almost all} of the four scenarios}: \scenei, \sceneii, \sceneiii \:and \sceneiv \:(Section \ref{subsec:models-list}), \textit{proving the efficacy of our technique}. We also see that for \oposumb \:benchmark, our models lag behind. However, \textit{this is expected}, as we highlight in Section \ref{subsec:benchmark-datasets}.}
    \label{tab:automatic-eval-oposum}
\end{table*}

\begin{table*}[t]
    \centering
    \resizebox{2\columnwidth}{!}{
        \begin{tabular}{*{10}c}
            \toprule
            Model Code & \amazonb & \amazonbr & \amazonbrdq & \oposumb & \oposumbr & \oposumbrdq & \flipkartb & \flipkartbr & \flipkartbrdq \\
            \midrule
            \tjmodel
             & \multirow{2}{*}{$\mathbf{88.78}$} & \multirow{2}{*}{$86.94$} & \multirow{2}{*}{$86.76$} & \multirow{2}{*}{$\mathbf{86.63}$} & \multirow{2}{*}{$86.96$} & \multirow{2}{*}{$86.95$} & \multirow{2}{*}{$-$} & \multirow{2}{*}{$-$} & \multirow{2}{*}{$-$} \\
             \cite{siledar-etal-2023-synthesize} &  &  &  &  &  &  &  &  &  \\
            \midrule
            \dpomodel & $86.45$ & $86.60^\dagger$ & $86.37^\dagger$ & $84.39$ & $87.35^*$ & $86.90$ & $83.75$ & $86.61$ & $85.40$ \\
            \supmodel & $87.79$ & $88.23^*$ & $87.76^*$ & $85.13$ & $88.59^*$ & $88.02^*$ & $84.21$ & $88.11$ & $86.40$ \\
            \naivemeanmodel & $87.95$ & $\underline{88.29}^*$ & $\underline{87.81}^*$ & $85.25$ & $88.96^*$ & $88.39^*$ & $\underline{84.32}$ & $\underline{88.29}$ & $\underline{86.52}$ \\
            \syntheticmodel & $87.81$ & $88.28^*$ & $87.74^*$ & $85.22$ & $\underline{89.08}^*$ & $\underline{88.45}^*$ & $84.27$ & $88.28$ & $86.49$ \\
            \inductivebiasmodel & $\underline{87.98}$ & $\mathbf{88.41}^*$ & $\mathbf{88.16}^*$ & $\underline{85.33}$ & $\mathbf{89.09}^*$ & $\mathbf{88.46}^*$ & $\mathbf{84.33}$ & $\mathbf{88.34}$ & $\mathbf{86.61}$ \\
            \bottomrule
        \end{tabular}
    }
    \caption{\textsc{BertScore} evaluation results on the $9$ benchmark datasets. We observe a similar trend as \textsc{Rouge} evaluations: SOTA is better than our models for the \amazonb \:and \oposumb \:benchmarks, which is expected (Section \ref{subsec:benchmark-datasets}). For the rest of the datasets, we see that our models are significantly better. We do not include SOTA results for Flipkart-based benchmarks, as \tjmodel \:only provide aspect-specific summaries for the same. $^*$ denotes gain is statistically significant compared to SOTA with significance level $1\%$, $^\dagger$ denotes gain is statistically significant  compared to SOTA with significance level $5\%$.}
    \label{tab:auto-eval-bertscore}
\end{table*}

\section{RLHF Training Pipeline}\label{asec:rlhf-training}

Using the trained reward model, we follow a similar training pipeline as \citet{bai-etal-2022-hh-rlhf-data, ouyang-etal-2022-instruct-gpt}, with a modification: \textit{Limited Trajectory Reinforcement Learning}. Computing the transformation $\Phi$ for each generation online (during training) is expensive, especially with limited compute resources. To circumvent this, we limit the trajectories that are explored by our policy, $\pi_\theta$. Specifically, we limit it to the \goodsum, \sbadsum \:and \vbadsum \:trajectories in the \propdata \:dataset. Having varying levels of quality in \propdata \:is of use here---it lets the model still explore trajectories of several quality. Thus, we have an offline experience buffer, with $\Phi$ precomputed, for $\pi_\theta$ learn from.

We use Proximal Policy Optimization (PPO) \cite{schulman-etal-2017-ppo} to train our model (Equation \ref{eqn:ppo-loss}). For each training step, we sample ($R, s, \Phi(R, s)$) tuples from \propdata. We use the trained \rmop \:to compute the reward for $s$ ($= \varphi_{op}\big(\Phi(R, s)\big)$). PPO uses this to update the log probability assigned by $\pi_\theta$. We parameterize $\pi_\theta$ using a Transformer model, which takes reviews as input, and generates an opinion summary.

\begin{align}\label{eqn:ppo-loss}
    \mathcal{L}_{PPO} = -\mathbb{E}&_{(R, s, \Phi(s))} \bigg[ \varphi_{op}(\Phi(R, s)) \nonumber\\ 
    &- \beta \log\Big(\dfrac{\pi^{RL}_\theta(s | R)}{\pi^{SFT}(s | R)}\Big) \bigg]
\end{align}

\section{Additional Automatic Evaluation Results}\label{asec:auto-eval-fo-benchmarks}

In addition to the Amazon-based benchmarks (Table \ref{tab:automatic-eval-amazon}), we also report results for Flipkart and Oposum+ based benchmarks (Tables \ref{tab:automatic-eval-flipkart} and \ref{tab:automatic-eval-oposum}). As before, we see that \inductivebiasmodel \:is almost always the winner.  As before, we draw similar conclusions for \scenei, \sceneii \:and \sceneiii: \inductivebiasmodel \:wins, further strengthening the conclusion that our methodology is effective. We also see that, inspite of the shortcomings of the \flipkartb \:benchmark, our models perform similar to the SOTA.

We also include \textsc{BertScore} evaluations for all the $9$ benchmarks in Table \ref{tab:auto-eval-bertscore}. We see similar trends as \textsc{Rouge} Evaluation: our models are significantly better than the SOTA in majority of the benchmarks.

For a qualitative understanding, we include generations from several models on a randomly picked sample from the \amazonb \:benchmark in Table \ref{tab:gen-example}.

\section{Details on the Benchmark Datasets}\label{asec:benchmark-problems}
In this section we discuss details about the benchmarks, such as the domain of the products, summary statistics and finally highlight some shortcomings in the \amazonb, \oposumb \:and \flipkartb \:datasets. \amazonb \:has reviews for $32$ products from $4$ domains: ``\textit{electronics}'', ``\textit{home \& kitchen}'', ``\textit{personal care}'', and ``\textit{clothing, shoes \& jewellery}''. \oposumb \:has reviews for $60$ products from $6$ domains: ``\textit{laptop bags}'', ``\textit{bluetooth headsets}'', ``\textit{boots}'', ``\textit{keyboards}'', ``\textit{television}'', and ``\textit{vacuums}''. \flipkartb \:has reviews for $147$ products from $3$ domains: ``\textit{laptops}'', ``\textit{mobiles}'', and ``\textit{tablets}''. Table \ref{tab:stats-benchmark-datasets} includes summary statistics for the benchmarks.

\begin{table}[h]
    \centering
    \resizebox{0.95\columnwidth}{!}{
        \begin{tabular}{*4c}
            \toprule
            Characteristic & \oposumb & \amazonb & \flipkartb \\
            \midrule
            \# domains & $6$ & $4$ & $3$ \\
            \# products & $60$ & $32$ & $147$ \\ \\
            \# reviews & \multirow{2}{*}{$10$} & \multirow{2}{*}{$8$} & \multirow{2}{*}{$10$} \\
            per product & & & \\ \\
            \# summaries & \multirow{2}{*}{$3$} & \multirow{2}{*}{$3$} & \multirow{2}{*}{$1$} \\
            per product & & & \\
            \bottomrule
        \end{tabular}
    }
    \caption{Statistics of the benchmark datasets. \oposumb \:represents the statistics of all \oposumb \:based benchmarks (\oposumb, \oposumbr \:and \oposumbrdq). Similar is the case for \amazonb \:and \flipkartb.}
    \label{tab:stats-benchmark-datasets}
\end{table}

Finally, now we highlight the shortcomings of the benchmark datasets in the rest of the discussion.

\noindent \textbf{\amazonb}: \citet{brazinskas-etal-2020-unsupervised} designed the test-set in such a way that the summary has to read like a review, for instance, summary would contain `\textit{I think the quality has come down over the years.}', instead of `\textit{Users think that quality has come down over years}'. Due to this writing style, the summaries read like reviews and are often in first person---high overlap would not necessarily mean a better summary, it would rather mean a better review.

\noindent \textbf{\flipkartb}: \citet{siledar-etal-2023-opinion-summ} generate this dataset by listing out the aspect-wise pros and cons presented within the reviews. We form an opinion summary by concatenating these pros and cons. Due to this, the summaries have frequent incoherent sentences. 

\noindent \textbf{\oposumb}: \citet{amplayo-etal-2021-aspect} create this benchmark by extracting sentences from the input reviews. Hence, this dataset has similar drawbacks as the \amazonb \:benchmark.

\vspace{0.5em}

\noindent\textbf{\amazonb}
\begin{MyIndentedList}
    \item
    \begin{MyIndentedList}
        \item \textit{Nice boots but run a bit narrow. They look great but I think the quality has come down over the years. Still comfortable but I wish they broke in easier. I recommend these for any lady who is patient and looking for comfort. }
    \end{MyIndentedList}
\end{MyIndentedList}

\noindent\textbf{\oposumb}
\begin{MyIndentedList}
    \item
    \begin{MyIndentedList}
        \item \textit{great product for the cost . very easy to use and compatible with all of my phones ! it holds a charge great , is light enough and fits perfectly in my ear . the sound quality is great , the style is very cool and the unit feels top quality . it would drop and reconnect every 10 seconds nobody could hear me i could n't get it to unpair from the phone , there 's apparently no noise-cancellation in these . the battery life is ... bizarre . cheap , plastic-y , and poor sound quality .}
    \end{MyIndentedList}
\end{MyIndentedList}

\textbf{\flipkartb}
\begin{MyIndentedList}
\item \textit{\textbf{Summary}}
    \begin{MyIndentedList}
        \item \textit{\textbf{Pros}}
        \begin{MyIndentedList}
            \item \textit{Design: The full-metal Infinix INBook X1 Core i3 has a top notch and premium design.}
            \item \textit{35.56 cm (14 inch) 1920 x 1080 Pixel Full HD IPS Display: 100\% sRGB with 300nits brightness ensures an excellent display.}
            \item \textit{Battery: Long-lasting battery. Gives around 8 hours of backup on normal usage.}
            \item \textit{Performance: The combination of Intel Core processor chip, high RAM size and sufficient storage capacity gives this laptop a high-speed performance experience.}
            \item \textit{Price: "Totally worth it in this price range.}
         \end{MyIndentedList} 
    \end{MyIndentedList}
    \begin{MyIndentedList}
        \item \textit{\textbf{Cons}}
        \begin{MyIndentedList}
            \item \textit{Charging: Some current leakage during charging. Sometimes the laptop won't charge.}
            \item \textit{Trackpad: Not upto the mark.}
         \end{MyIndentedList} 
    \end{MyIndentedList}
    \begin{MyIndentedList}
        \item \textit{\textbf{Verdict}: This laptop comes with a i3 10th gen dual core processor which is suitable for normal tasks like web browsing, online classes and watching movies. Not recommended as a gaming laptop.}
    \end{MyIndentedList}
    \begin{MyIndentedList}
        \item \textit{\textbf{Additional Information}: Can handle video editing and expandable SSD.}
    \end{MyIndentedList}
\end{MyIndentedList}

\section{Implementation Details}\label{asec:implementation-details}
We use BART-Large \cite{lewis-etal-2020-bart} as our policy ($\pi_\theta$) in all of the models. We do this to have a fair comparison with the state-of-the-art in Opinion Summarization. We use AdamW Optimizer to train the models, with a weight decay of $0.05$. We use a {\tt cosine} learning rate scheduler. We run a hyperparameter sweep on {\tt batch} {\tt size}, {\tt learning} {\tt rate}, and {\tt learning} {\tt rate} {\tt warmup}. We include the possible values for the sweep in Table \ref{tab:implementation-details-policy}. We train all of our models using $2\times$ A$100$ GPUs ($80$GB)

\begin{table}[h]
    \centering
    \begin{tabular}{*2c}
        \toprule
         Hyperparameter & Values \\
         \midrule
         {\tt batch} {\tt size} & [$64$, $128$, $256$] \\
         {\tt learning} {\tt rate} & $\sim\mathcal{U}(5e^{-6}, 5e^{-5})$ \\ 
         {\tt learning} {\tt rate} {\tt warmup} & $\sim\mathcal{U}(0.2, 0.4)$\\
         \bottomrule
    \end{tabular}
    \caption{Possible Values for Hyperparameters. For {\tt learning} {\tt rate} {\tt warmup}, we sample the fraction of total steps the learning should be warmed up. For example, if the {\tt learning} {\tt rate} {\tt warmup} is $0.2$, it means that the learning rate will have a linear warmup for $20\%$ of the total training steps.}
    \label{tab:implementation-details-policy}
\end{table}

For the reward model, \rmop, we use a Feed Forward Network for the Policy Model. We use AdamW Optimizer to train the models, with a weight decay of $0.05$. As before, we run a hyperparameter sweep on {\tt batch} {\tt size}, {\tt learning} {\tt rate}, and {\tt learning} {\tt rate} {\tt warmup}. Table \ref{tab:implementation-details-reward} includes details on the hyperparameters.

\begin{table}[h]
    \centering
    \begin{tabular}{*2c}
        \toprule
         Hyperparameter & Values \\
         \midrule
         {\tt batch} {\tt size} & [$32$, $64$, $128$] \\
         {\tt learning} {\tt rate} & $\sim\mathcal{U}(5e^{-3}, 1e^{-1})$ \\ 
         \bottomrule
    \end{tabular}
    \caption{Possible Values for Hyperparameters for the Reward Model. For {\tt learning} {\tt rate} {\tt warmup}, we sample the fraction of total steps the learning should be warmed up. For example, if the {\tt learning} {\tt rate} {\tt warmup} is $0.2$, it means that the learning rate will have a linear warmup for $20\%$ of the total training steps.}
    \label{tab:implementation-details-reward}
\end{table}

\section{Generated Summary Lengths}\label{asec:gen-summary-length}
We analyze the generation lengths of the models, and the ground truth summary. Table \ref{tab:gen-summary-length} lists the summary lengths. We see that the\dpomodel \:model generates very verbose summary. Additionally, we also see that the \inductivebiasmodel \:model generates very concise summaries.

\begin{table}[h]
    \centering
    \small
    \resizebox{\columnwidth}{!}{%
    \begin{tabular}{lccc}
    \toprule
     Model & \amazonb & \oposumb & \flipkartb \\
     \midrule
    Ground-Truth & $60.65$ & $85.86$ & $129.91$\\
 \naivemeanmodel & $91.09$ & $114.67$ & $75.48$ \\ 
 \syntheticmodel & $80.31$ & $115.37$ & $71.11$\\ 
 \tjmodel & $55.84$ & $62.93$ & - \\
 \inductivebiasmodel & $81.62$ & $88.63$ & $73.57$ \\
 \supmodel & $81.31$ & $117.03$ & $74.56$\\
 \dpomodel & $138.50$ & $141.50$ & $131.40$\\
    \bottomrule
    \end{tabular}
    }
\caption{Generation Length Statistics: number of words in summaries. We use \textsc{Nltk} to tokenize the text.}
\label{tab:gen-summary-length}
\end{table}

\begin{table*}[t]
\centering
\resizebox{2\columnwidth}{!}{
\begin{tabular}{p{15cm}}
\toprule
\textbf{\textit{Good}}:
The users have shared their positive experiences with the protective covers for Macbooks. They appreciate the ease of application, vibrant colors, quick shipping, and great quality. Some have mentioned the covers fit their Macbooks perfectly, while others have shared their disappointment when they realized it did not fit their specific model. A few users have noted the covers add a personal touch and pizzaz to their devices. However, some users have reported issues with the covers not fitting their Macbooks or falling off, leaving them feeling disappointed and frustrated.
 \\
\midrule

\textbf{\textit{Slightly bad}}:
Some users have reported positive experiences with the Macbook cover, praising its protective qualities, easy application, and vibrant colors. Others, however, have encountered issues with it not fitting properly on their devices or being returned due to size mismatches. Overall, the product has been described as cute, great quality, and worth the investment, though some buyers have experienced disappointment with its size compatibility and easy detachment.
 \\
\midrule

\textbf{\textit{Very Bad}}:
This is a terrible product for people with old Macbooks, it doesn't fit at all! And don't even bother looking at the description carefully before buying.
 \\
\bottomrule
\end{tabular}
}
\caption{Example summaries from \propdata.}
\label{tab:ex-dataset}
\end{table*}

\section{Details on \propdata}\label{asec:generated-data-appendix}
Here we provide more details on the generated \propdata \:dataset. Table \ref{tab:stats-data-gen} includes summary statistics of the generated dataset. We include an example from \propdata \:dataset (Table \ref{tab:ex-dataset}). We show one sample from \goodsum, \sbadsum \:and \vbadsum \:quality each. We do not include the reviews for conciseness. However, we incluce the salient aspects of the reviews. The reviews talk about the following things:

\begin{enumerate}
    \item Great price, Nice looking / Good color, Good utility / Good protection, Quick shipping, Nice fitting, Good accessibility of the laptop while the cover is on, Good finish quality.
    \item Not a good fit for older macbooks, Broken / Unusable for the original ($\sim2009$ - $2010$) white macbooks, cannot return return.
\end{enumerate}

\goodsum \:manages to discuss all of these things---it was able to detect that the cover does not fit specific models (highlighted in blue). It also detects that a few users like the ``personal touch addition'' factor of the cover.

\vbadsum \:is also a good representation of a bad summary---it totally ignores the ``positive'' aspects of the product and presents the ``negative'' aspects only. It fails at Aspect Coverage, Opinion Faithfulness and Opinion Coverage. 

\sbadsum \:maintains almost a similar quality as the Good one. However, it fails to draw out certain aspects, such as ``pizzaz'', ``personal touch addition'', etc.

\section{Statistics of the \opinpref \:dataset}\label{asec:opinpref-stats}
We look at the summary statistics for the \opinpref \:dataset. Table \ref{tab:stats-opinpref-dataset}. We see that, interestingly, annotators prefer longer summaries---this is because these summaries contain more specifics and details from the reviews.

\begin{table}[h]
    \centering
    \resizebox{0.95\columnwidth}{!}{
        \begin{tabular}{*2c}
                       \toprule
            Characteristic & Value\\
            \midrule

    \# words in reviews & $641.21$ \\
    \# reviews & $13.08$ \\
    \# words in summaries & $73.16$ \\
    \# words in preferred summaries & $85.41$ \\
    \# words in unpreferred summaries & $66.91$ \\
    \bottomrule
    \end{tabular}
    }
    \caption{Statistics of the \opinpref \:dataset. We use \textsc{Nltk} to tokenize the text.}
    \label{tab:stats-opinpref-dataset}
\end{table}

\begin{table}[h]
    \centering
    \resizebox{0.95\columnwidth}{!}{
        \begin{tabular}{*4c}
            \toprule
            Split & Characteristic & $\mu$ & $\sigma$\\
            \midrule
            \multirow{4}{*}{\trainset} & \# reviews & $13.24$ & $10.07$ \\
            & \# summaries & $8.90$ & $0.34$ \\
            & \# words in review & $49.0$ & $10.78$ \\
            & \# words in summary & $78.28$ & $34.45$ \\
            \midrule
            \multirow{4}{*}{\validset} & \# reviews & $10.53$ & $6.80$ \\
            & \# summaries & $8.98$ & $0.16$ \\
            & \# words in review & $48.65$ & $10.63$ \\
            & \# words in summary & $74.26$ & $34.27$ \\
            \bottomrule
        \end{tabular}
    }
    \caption{Statistics of \propdata \:dataset. We use \textsc{Nltk} to tokenize the text.}
    \label{tab:stats-data-gen}
\end{table}

\begin{table*}[t]
\centering
\resizebox{2\columnwidth}{!}{
\begin{tabular}{p{15cm}}
\toprule
\textbf{\textit{Reviews}}: i really like these boots. they can be a beast to get on, like any boot fit to last; once on, they are incredibly comfortable. I have had them for 4 years and they still look great - worn in but not worn out...

What great boots! They do take some breaking in - like all real leather does. Run maybe a tad small in the general spectrum, but hey, remember the break in period. Look fabulous, will last me for the next twenty years.... They bridge the fashion vs. function quandary beautifully.

Every woman needs a pair of these Frye boots! Fit is true to normal size. I wear a standard 9. These fit in the shaft for any ladies who have bigger calves as I do. You have to fight to get them on first few times and they wear stiff until fully broken in. Worth the money and break in time.

Boots excellent quality, beautiful skin, a modern model. But! they are very narrow! This is the second pair of shoes is a brand and my 8.5 size fits me perfectly, even he little more than usual. But these are very narrow! go to a cobbler, let him try to stretch. Be careful!

bought these for the wife to have something to kick around in and keep up with the kids, but she doesn't like them. Her heel lifts with every step, and they're stiff and heavy. Not an everyday casual shoe.

Don't get me wrong, they're really good boots, but they used to make them better. My last pair from a few years ago were much higher quality. Thicker sole, better full grain leather, now they only make the black in 'grease', which stains all my socks black. Keeping them, but still disappointed in Frye.

i bought it just for the heavy snow cuz my UGG got wet through.even though it is a little expensive, but the boots are nice and it was shipped to me for only two days.!!!!!! i love it

They were comfortable the moment I put them on. The sizes are exact. The color the way they are made are what I expect from Frye which I've been wearing since the 70's. With a 20year break from not wearing Frye it was a pleasant surprise the quality has stood the test of time. \\
\midrule
\textbf{\textit{\supmodel}}: These Frye boots are a popular choice for women due to their {\color{olive} comfort} and {\color{olive} durability}. However, they can be a bit challenging to put on and {\color{olive} require a break-in period}. Some users have reported that they run small in size and are narrow, requiring a {\color{olive} cobbler's assistance for a proper fit}. The {\color{olive} leather quality is generally good}, but some users have noted that the {\color{red} leather may not last as long as they had hoped}.\\
\midrule
\textbf{\textit{\naivemeanmodel}}: Users generally appreciate the {\color{olive} comfort} and {\color{olive} durability} of the Frye boots, with some mentioning the need for a {\color{olive} break-in period}. {\color{olive} Sizes run true} to normal, but some users find them to be a {\color{olive} bit narrow}. The boots are praised for their {\color{olive} stylish appearance} and ability to {\color{olive} bridge the fashion and function quandary}. However, some users have noted that they take {\color{red} some time to break in} and are {\color{olive} not suitable for everyday casual wear}.\\
\midrule
\textbf{\textit{\syntheticmodel}}: These Frye boots are praised for their {\color{olive} comfort} and {\color{olive} durability}, with some users noting the need for a {\color{olive} break-in period}. However, some users have noted that they run small in size and require stretching, while others find them {\color{red} narrow and stiff}. The {\color{olive} leather is described as beautiful and of good quality}, with a {\color{olive} modern model}. Some users have {\color{red} had them for several years} and have not worn them out.\\
\midrule
\textbf{\textit{\inductivebiasmodel}}: These Frye boots are praised for their {\color{olive} comfort} and {\color{olive} durability}, with users expressing satisfaction with their {\color{olive} fit} and {\color{olive} appearance}. However, some users have noted that they require a {\color{olive} break-in period} and can be a {\color{olive} bit narrow}. The {\color{olive} leather quality is praised} for its durability and {\color{olive} modern model}, with some users expressing {\color{olive} disappointment with the lack of improved quality in recent years}.\\
\bottomrule
\end{tabular}
}
\caption{Example generation (randomly sampled) for some input reviews from all the models. {\color{olive} Olive} implies faithful/correct generation, while {\color{red} red} indicates hallucinated text, or repetition. We see that only \inductivebiasmodel \:is free from {\color{red} red} text. The model closest in performance to \inductivebiasmodel, the \naivemeanmodel \:model, misses out on two aspects: {\tt leather-quality} and {\tt quality-degradation}. \inductivebiasmodel \:covers both, while being concise. We do not include \dpomodel \:model in this comparison, as it was too verbose.}
\label{tab:gen-example}
\end{table*}

\section{Annotator Details}\label{asec:annotator-details}

We include two disjoint sets of annotators in our work---first for creation of \opinpref \:($3$ annotators), second for human evaluation \:($3$ annotators). For both annotations, we use domain experts. The domain experts are NLP researchers (age group: $24-30$) who have worked in Opinion Summarization for a long time, with publication experience (in \textit{A}/\textit{A}$^*$ conferences). The domain experts for human evaluation also have a similar profile. The annotators have been paid generously, based on the standard annotation rates in the geographical location.

\section{All Evaluation Results}\label{asec:all-eval-results}
We include all of the evaluation results in this section. In Tables \ref{tab:human-eval-amazon}, \ref{tab:gpt4-evaluation-amazon}, \ref{tab:gpt4-evaluation-flipkart} and \ref{tab:gpt4-evaluation-flipkart} we include pairwise comparison results, in a {\tt win}/{\tt tie}/{\tt loss} format. We also include results on evaluation on how the models perform on the domain features in Tables \ref{tab:intrinsic-eval-amazon}, \ref{tab:intrinsic-eval-flipkart} and \ref{tab:intrinsic-eval-oposum}.

\begin{table*}[t]
    \centering
    \resizebox{1.9\columnwidth}{!}{
        \begin{tabular}{l*5c}
            \toprule
            $\cdot$ & \supmodel & \naivemeanmodel & \syntheticmodel & \inductivebiasmodel & \tjmodel \\
            \midrule
            \naivemeanmodel & $0.50/0.06/0.38$				 \\
            \syntheticmodel & $0.44/0.12/0.44$	& $0.40/0.09/0.5$	\\
            \inductivebiasmodel & $\mathbf{0.56}/0.09/0.28$ &	$\mathbf{0.46}/0.18/0.31$	 & $\mathbf{0.56}/0.12/0.28$	\\
            \tjmodel &
            $0.31/0.28/0.38$ &	$0.25/0.12/0.56$ & $0.25/0.21/0.5$  & $0.25/0.06/\mathbf{0.68}$	\\
            Ground-Truth	& $0.46/0.06/0.48$ & 	$0.31/0.18/0.44$ &	$0.40/0.15/0.40$	& $0.28/0.09/\mathbf{0.59}$ & 	$0.5/0.09/0.38$ \\
                \bottomrule
        \end{tabular}
    }
    \caption{Pairwise Win/Tie/Loss Results for all models in Human Evaluation for \amazonb \:benchmark. We format the data as: {\tt win}/{\tt tie}/{\tt loss}, {\tt win} specifies how many time the \textit{row} won over the \textit{column}.}
    \label{tab:human-eval-amazon}
\end{table*}

\begin{table*}[t]
    \centering
    \resizebox{1.9\columnwidth}{!}{
        \begin{tabular}{l*5c}
            \toprule
            $\cdot$ & \supmodel & \naivemeanmodel & \syntheticmodel & \inductivebiasmodel & \tjmodel \\
            \midrule
            \naivemeanmodel & $0.63/0.12/0.25$					 \\
            \syntheticmodel & $0.59/0.12/0.28$	& $0.5/0.06/0.44$ \\
            \inductivebiasmodel & $\mathbf{0.62}/0.12/0.25$ &	$\mathbf{0.46}/0.09/0.44$	 & $ \mathbf{0.5}/0.06/0.44$	\\
            \tjmodel & $0.06/0.03/0.9$	 &	$0.09/0.0/0.90$	 & $0.12/0.09/0.78$	&  $0.06/0.0/\mathbf{0.93}$\\
            ground-truth	& $0.12/0.06/0.81$ &	$0.09/0.06/0.84$	& $0.16/0.06/0.78$	& $0.09/0.0/\mathbf{0.90}$	& $0.68/0.09/0.22$ \\
            \bottomrule
        \end{tabular}
    }
    \caption{Pairwise Win/Tie/Loss Results for all models in \gpt{4} Evaluation for \amazonb \:benchmark. We format the data as: {\tt win}/{\tt tie}/{\tt loss}, {\tt win} specifies how many time the \textit{row} won over the \textit{column}.}
    \label{tab:gpt4-evaluation-amazon}

\end{table*}

\begin{table*}[t]
    \centering
    \resizebox{1.75\columnwidth}{!}{
        \begin{tabular}{l*4c}
            \toprule
            $\cdot$ & \supmodel & \naivemeanmodel & \syntheticmodel & \inductivebiasmodel \\
            \midrule
            \naivemeanmodel & $0.57/0.12/0.30$ \\
            \syntheticmodel & $0.57/0.06/0.36$ & $0.52/0.12/0.36$		\\
            \inductivebiasmodel & $\mathbf{0.63}/0.12/0.25$ & $\mathbf{0.54}/0.16/0.30$ &	$\mathbf{0.57}/0.08/0.34$ \\
            Ground-Truth & $0.10/0.06/0.84$ & $0.06/0.01/0.92$	& $0.07/0.01/0.91$	& $0.06/0.02/\mathbf{0.91}$	\\
            \bottomrule
        \end{tabular}
    }
    \caption{Pairwise Win/Tie/Loss Results for all models in \gpt{4} Evaluation for \flipkartb \:benchmark. We format the data as: {\tt win}/{\tt tie}/{\tt loss}, {\tt win} specifies how many time the \textit{row} won over the \textit{column}.}
    \label{tab:gpt4-evaluation-flipkart}
\end{table*}

\begin{table*}[t]
    \centering
    \resizebox{1.9\columnwidth}{!}{
        \begin{tabular}{l*5c}
            \toprule
            $\cdot$ & \supmodel & \naivemeanmodel & \syntheticmodel & \inductivebiasmodel & \tjmodel \\
            \midrule
            \naivemeanmodel & $0.56/0.03/0.4$	 \\
            \syntheticmodel & $0.5/0.16/0.34$ &	$0.46/0.1/0.44$	\\
            \inductivebiasmodel & $\mathbf{0.66}/0.0/0.33$	& $\mathbf{0.46}/0.1/0.44$ & $\mathbf{0.56}/0.06/0.36$		\\
            \tjmodel & $0.1/0.06/0.83$ & $0.06/0.03/0.9$	& $0.03/0.03/0.93$	& $0.03/0.03/\mathbf{0.93}$		\\
            Ground-Truth & $0.13/0.13/0.73$	& $0.1/0.033/0.8666$	& $0.06/0.06/0.86$ & $0.06/0.06/\mathbf{0.86}$	& $0.7/0.1/0.2$ \\
            \bottomrule
        \end{tabular}
    }
    \caption{Pairwise Win/Tie/Loss Results for all models in \gpt{4} Evaluation for \oposumb \:benchmark. We format the data as: {\tt win}/{\tt tie}/{\tt loss}.}
    \label{tab:gpt4-evaluation-oposum}
\end{table*}

\begin{table*}[t]
    \centering
    \resizebox{2\columnwidth}{!}{
        \begin{tabular}{l*7c}
            \toprule
            $\cdot$ & AC & OPF & OPC & CC & RL & HL & LC \\
            \midrule
            \supmodel & $3.43 \pm 0.20$ & $3.71 \pm 0.37$ & $3.67 \pm 0.26$ & $3.79 \pm 0.31$ & $4.04 \pm 0.37$ & $3.89 \pm 0.39$ & $4.55 \pm 0.35$ \\  
            \naivemeanmodel & $3.56 \pm 0.22$ & $3.91 \pm 0.50$ & $3.76 \pm 0.38$ & $3.89 \pm 0.36$ & $4.04 \pm 0.48$ & $3.99 \pm 0.48$ & $4.60 \pm 0.27$ \\
            \syntheticmodel & $3.55 \pm 0.40$ & $3.87 \pm 0.71$ & $3.71 \pm 0.43$ & $3.94 \pm 0.50$ & $4.04 \pm 0.61$ & $3.94 \pm 0.68$ & $4.38 \pm 0.92$ \\
            \inductivebiasmodel & $\mathbf{3.60} \pm \mathbf{0.17}$ & $\mathbf{3.95} \pm \mathbf{0.40}$ & $\mathbf{3.85} \pm \mathbf{0.25}$ & $3.99 \pm 0.35$ & $4.06 \pm 0.34$ & $\mathbf{4.07} \pm \mathbf{0.43}$ & $\mathbf{4.65 }\pm \mathbf{0.32}$ \\
            \tjmodel & $3.34 \pm 0.68$ & $3.92 \pm 0.79$ & $3.70 \pm 0.54$ & $\mathbf{4.0} \pm \mathbf{0.50}$ & $\mathbf{4.08} \pm \mathbf{0.72}$ & $3.87 \pm 1.08$ & $4.05 \pm 1.31$ \\
            Ground-Truth & $3.55 \pm 0.50$ & $3.93 \pm 0.46$ & $3.56 \pm 0.31$ & $4.08 \pm 0.32$ & $4.04 \pm 0.46$ & $3.81 \pm 0.86$ & $4.40 \pm 0.45$ \\
\bottomrule
        \end{tabular}
    }
    \caption{Intrisic Evaluation results on the \amazonb \:benchmark for all the models. Legend: AC: \maspcov, OPF: \mopfaith, OPC: \mopcov, CC: \mconcise, RE: \mrelevance, HL: \mhallucination, LC: \mlangcorr.}
    \label{tab:intrinsic-eval-amazon}
\end{table*}

\begin{table*}[t]
    \centering
    \resizebox{2\columnwidth}{!}{
        \begin{tabular}{l*7c}
            \toprule
            $\cdot$ & AC & OPF & OPC & CC & RL & HL & LC \\
            \midrule
            \supmodel & $3.61 \pm 0.22$ & $4.10 \pm 0.39$ & $3.84 \pm 0.33$ & $4.04 \pm 0.28$ & $4.21 \pm 0.31$ & $4.19 \pm 0.42$ & $4.53 \pm 0.27$ \\  
            \naivemeanmodel & $3.56 \pm 0.21$ & $4.13 \pm 0.41$ & $3.84 \pm 0.34$ & $4.0 \pm 0.32$ & $\mathbf{4.31} \pm \mathbf{0.36}$ & $4.26 \pm 0.34$ & $4.54 \pm 0.39$ \\
            \syntheticmodel & $3.56 \pm 0.25$ & $4.09 \pm 0.40$ & $3.79 \pm 0.32$ & $\mathbf{4.02} \pm \mathbf{0.30}$ & $4.19 \pm 0.34$ & $4.19 \pm 0.36$ & $4.53 \pm 0.29$ \\
            \inductivebiasmodel & $\mathbf{3.63} \pm \mathbf{0.20}$ & $\mathbf{4.22} \pm \mathbf{0.39}$ & $\mathbf{3.85} \pm \mathbf{0.30}$ & $4.01 \pm 0.28$ & $4.26 \pm 0.29$ & $\mathbf{4.33} \pm \mathbf{0.45}$ & $\mathbf{4.61} \pm \mathbf{0.29}$ \\
            Ground-Truth & $3.59 \pm 0.15$ & $3.88 \pm 0.53$ & $3.68 \pm 0.27$ & $4.02 \pm 0.28$ & $3.87 \pm 0.59$ & $3.67 \pm 0.78$ & $4.35 \pm 0.44$ \\
            \bottomrule
        \end{tabular}
    }
    \caption{Intrisic Evaluation results on the \flipkartb \:benchmark for all the models. Legend: AC: \maspcov, OPF: \mopfaith, OPC: \mopcov, CC: \mconcise, RE: \mrelevance, HL: \mhallucination, LC: \mlangcorr.}
    \label{tab:intrinsic-eval-flipkart}
\end{table*}

\begin{table*}[t]
    \centering
    \resizebox{2\columnwidth}{!}{
        \begin{tabular}{l*7c}
            \toprule
            $\cdot$ & AC & OPF & OPC & CC & RL & HL & LC \\
            \midrule
            \supmodel & $3.47 \pm 0.14$ & $3.38 \pm 0.26$ & $3.49 \pm 0.06$ & $\mathbf{3.64} \pm \mathbf{0.19}$ & $\mathbf{3.81} \pm \mathbf{0.26}$ & $3.22 \pm 0.56$ & $3.96 \pm 0.32$ \\  
            \naivemeanmodel & $3.49 \pm 0.05$ & $3.48 \pm 0.06$ & $3.5 \pm 0.0$ & $3.56 \pm 0.13$ & $3.66 \pm 0.22$ & $3.52 \pm 0.33$ & $4.1 \pm 0.33$ \\
            \syntheticmodel & $3.50 \pm 0.03$ & $3.41 \pm 0.26$ & $3.5 \pm 0.0$ & $3.63 \pm 0.24$ & $3.62 \pm 0.20$ & $3.32 \pm 0.63$ & $\mathbf{4.03} \pm \mathbf{0.38}$ \\
            \inductivebiasmodel & $\mathbf{3.54} \pm \mathbf{0.22}$ & $\mathbf{3.50} \pm \mathbf{0.06}$ & $\mathbf{3.57} \pm \mathbf{0.06}$ & $3.62 \pm 0.19$ & $3.65 \pm 0.23$ & $\mathbf{3.68} \pm \mathbf{0.36}$ & $4.0 \pm 0.29$ \\
            \tjmodel & $3.39 \pm 0.3$ & $3.46 \pm 0.45$ & $3.49 \pm 0.28$ & $3.61 \pm 0.40$ & $3.58 \pm 0.82$ & $3.43 \pm 0.92$ & $3.79 \pm 1.18$ \\
            Ground-Truth & $3.42 \pm 0.22$ & $3.475 \pm 0.28$ & $3.5 \pm 0.0$ & $3.57 \pm 0.16$ & $3.49 \pm 0.28$ & $3.21 \pm 0.48$ & $3.56 \pm 0.23$ \\
            \bottomrule
        \end{tabular}
        }
    \caption{Intrisic Evaluation results on the \oposumb \:benchmark for all the models. Legend: AC: \maspcov, OPF: \mopfaith, OPC: \mopcov, CC: \mconcise, RE: \mrelevance, HL: \mhallucination, LC: \mlangcorr.}
    \label{tab:intrinsic-eval-oposum}
\end{table*}










\end{document}